\documentclass[10pt,twocolumn,letterpaper]{article}

\usepackage[pagenumbers]{wacv} 

\definecolor{wacvblue}{rgb}{0.21,0.49,0.74}
\usepackage[pagebackref,breaklinks,colorlinks,allcolors=wacvblue]{hyperref}
\usepackage{pgfplots}
\usepackage{xcolor}
\usepackage{amsmath}
\usepackage{float} 

\usepgfplotslibrary{groupplots}
\pgfplotsset{compat=1.18}

\definecolor{fifacolor}{HTML}{5C677D}
\definecolor{visualcolor}{HTML}{3A7CA5}
\definecolor{avacolor}{HTML}{3A9D7A}
\definecolor{moecolor}{HTML}{D08C32}
\definecolor{saltcolor}{HTML}{B23A48}
\usepackage{graphicx}
\usepackage{tikz}
\usetikzlibrary{arrows.meta,shapes.geometric,calc}
\def\wacvPaperID{11} 
\def\confName{WACV}
\def\confYear{2027}

\title{SALT: State-Aligned Low-Rank Transfer for Single-Temporal Cross-Dataset Structural Localization in Remote Sensing}

\author{
Aggelos Psiris$^{1}$ \quad
Georgios Angelidis$^{1}$ \quad
Giorgos Sfikas$^{2}$ \quad
Maria Vakalopoulou$^{3,6}$ \\
Thomas Lagkas$^{4}$ \quad
Panagiotis Sarigiannidis$^{5}$ \quad
Georgios Th. Papadopoulos$^{1,6}$ \\[0.5em]
$^{1}$Harokopio University of Athens \qquad
$^{2}$University of West Attica \\
$^{3}$CentraleSup\'elec, Universit\'e Paris-Saclay \\
$^{4}$Democritus University of Thrace \qquad
$^{5}$University of Western Macedonia \\
$^{6}$Archimedes Research Unit, Athena Research Center \\[0.3em]
{\tt\small $^{1}$\{aggelospsiris, gangelidis, G.Th.Papadopoulos\}@hua.gr} \\
{\tt\small $^{2}$gsfikas@uniwa.gr \quad
$^{3}$maria.vakalopoulou@centralesupelec.fr} \\
{\tt\small $^{4}$tlagkas@cs.duth.gr \quad
$^{5}$psarigiannidis@uowm.gr}
}

\begin{document}
\maketitle

\begin{abstract}
Change detection in remote sensing conventionally relies on spatially aligned bi-temporal imagery, which may be unavailable or unreliable in practical scenarios. Recent single-temporal approaches address this limitation by learning structural evidence, such as building regions, from a labeled source dataset and transferring it to unseen target imagery using only the current image. However, adaptation to the source dataset can over-specialize pretrained features and reduce their cross-dataset generalization, a risk that this paper argues is higher when the feature adaptation is parameterized independently of the states used for prediction. To address this issue, the current study introduces SALT, a State-Aligned Low-Rank Transfer framework. SALT keeps a DINOv3 backbone frozen and learns negative and positive structural states that both score each patch and, through State-Aligned Low-Rank Adaptation (SALA), span the rank-2 subspace in which patch features are corrected. Directional State Alignment supervises these corrections to raise the evidence of positive patches and lower that of negative ones, while Geometry Preservation limits changes to the pairwise relations among pretrained patch features. Experiments on LEVIR-CD and the WHU aerial building dataset show that, at the pixel level and in both transfer directions, SALT outperforms four baselines reproduced under a common source-only protocol,  improving over the strongest reproduced baselines by $9.45$ and $5.79$ points in pixel-level AP and by $27.69$ and $14.25$ points in IoU (LEVIR-CD$\rightarrow$WHU and WHU$\rightarrow$LEVIR-CD, respectively), with only $0.068$M trainable parameters and $40.2$ ms per image. In order to facilitate reproducibility and further research, the source code of SALT is available at https://github.com/gthpapadopoulos/SALT

\end{abstract}  

\section{Introduction}
\label{sec:introduction}

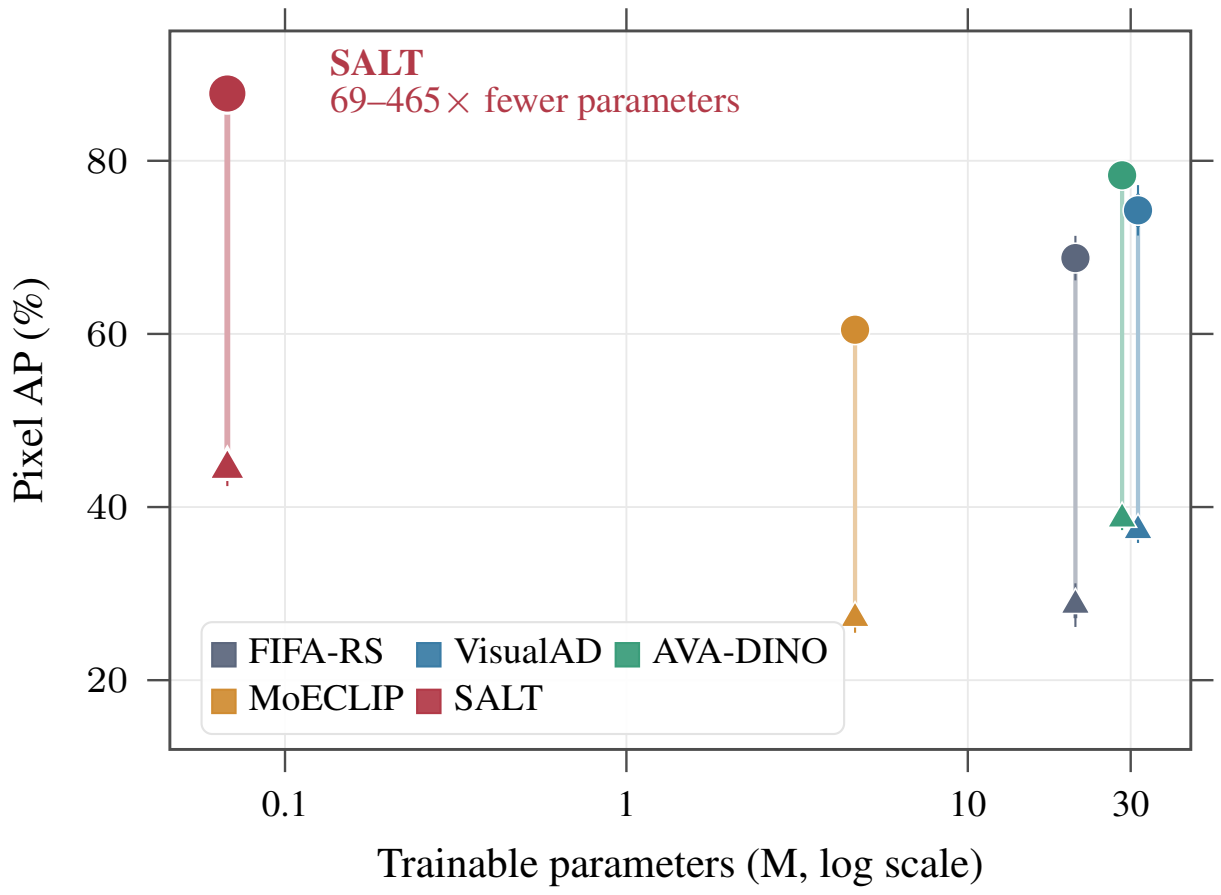
\begin{figure}[t]
\centering

\pgfplotsset{
    saltdot/.style={
        only marks,
        mark=*,
        mark size=2.8pt,
        error bars/y dir=both,
        error bars/y explicit,
        error bars/error bar style={line width=0.7pt},
        error bars/error mark options={mark size=1.65pt}
    },
    salttriangle/.style={
        only marks,
        mark=triangle*,
        mark size=3.15pt,
        error bars/y dir=both,
        error bars/y explicit,
        error bars/error bar style={line width=0.7pt},
        error bars/error mark options={mark size=1.65pt}
    }
}

\begin{tikzpicture}
\begin{axis}[
    width=\columnwidth,
    height=0.76\columnwidth,
    xmode=log,
    xmin=0.046,
    xmax=45,
    ymin=12,
    ymax=95,
    xlabel={Trainable parameters (M, log scale)},
    ylabel={Pixel AP (\%)},
    xlabel style={font=\footnotesize,yshift=1pt},
    ylabel style={font=\footnotesize},
    tick label style={font=\scriptsize},
    xtick={0.1,1,10,30},
    xticklabels={0.1,1,10,30},
    ytick={20,40,60,80},
    tick align=outside,
    axis line style={
        draw=black!70,
        line width=0.55pt
    },
    tick style={
        draw=black!70,
        line width=0.45pt
    },
    xmajorgrids=true,
    ymajorgrids=true,
    major grid style={
        draw=gray!17,
        line width=0.35pt
    },
    legend columns=3,
    legend style={
        at={(0.03,0.02)},
        anchor=south west,
        font=\scriptsize,
        cells={anchor=west},
        draw=gray!22,
        fill=white,
        fill opacity=0.93,
        text opacity=1,
        rounded corners=2pt,
        inner sep=2pt,
        row sep=0.5pt,
        column sep=1.5pt,
        nodes={inner sep=1pt}
    },
    clip=false
]


\draw[draw=moecolor!45,line width=0.8pt]
    (axis cs:4.674560,27.1420) --
    (axis cs:4.674560,60.4849);

\draw[draw=fifacolor!45,line width=0.8pt]
    (axis cs:20.665,28.6685) --
    (axis cs:20.665,68.7595);

\draw[draw=visualcolor!45,line width=0.8pt]
    (axis cs:31.520788,37.2907) --
    (axis cs:31.520788,74.2757);

\draw[draw=avacolor!45,line width=0.8pt]
    (axis cs:28.311552,38.6280) --
    (axis cs:28.311552,78.3216);

\draw[draw=saltcolor!45,line width=1.1pt]
    (axis cs:0.067778,44.4188) --
    (axis cs:0.067778,87.7706);


\addplot+[
    forget plot,
    saltdot,
    color=fifacolor,
    mark options={
        fill=fifacolor,
        draw=white,
        line width=0.3pt
    }
] coordinates {
    (20.665,68.7595) +- (0,1.5585)
};

\addplot+[
    forget plot,
    saltdot,
    color=visualcolor,
    mark options={
        fill=visualcolor,
        draw=white,
        line width=0.3pt
    }
] coordinates {
    (31.520788,74.2757) +- (0,1.9004)
};

\addplot+[
    forget plot,
    saltdot,
    color=avacolor,
    mark options={
        fill=avacolor,
        draw=white,
        line width=0.3pt
    }
] coordinates {
    (28.311552,78.3216) +- (0,0.0867)
};

\addplot+[
    forget plot,
    saltdot,
    color=moecolor,
    mark options={
        fill=moecolor,
        draw=white,
        line width=0.3pt
    }
] coordinates {
    (4.674560,60.4849) +- (0,0.7226)
};

\addplot+[
    forget plot,
    saltdot,
    mark size=3.7pt,
    color=saltcolor,
    mark options={
        fill=saltcolor,
        draw=white,
        line width=0.6pt
    }
] coordinates {
    (0.067778,87.7706) +- (0,0.2074)
};


\addplot+[
    forget plot,
    salttriangle,
    color=fifacolor,
    mark options={
        fill=fifacolor,
        draw=white,
        line width=0.3pt
    }
] coordinates {
    (20.665,28.6685) +- (0,1.5076)
};

\addplot+[
    forget plot,
    salttriangle,
    color=visualcolor,
    mark options={
        fill=visualcolor,
        draw=white,
        line width=0.3pt
    }
] coordinates {
    (31.520788,37.2907) +- (0,0.4250)
};

\addplot+[
    forget plot,
    salttriangle,
    color=avacolor,
    mark options={
        fill=avacolor,
        draw=white,
        line width=0.3pt
    }
] coordinates {
    (28.311552,38.6280) +- (0,0.2757)
};

\addplot+[
    forget plot,
    salttriangle,
    color=moecolor,
    mark options={
        fill=moecolor,
        draw=white,
        line width=0.3pt
    }
] coordinates {
    (4.674560,27.1420) +- (0,0.6505)
};

\addplot+[
    forget plot,
    salttriangle,
    mark size=4.0pt,
    color=saltcolor,
    mark options={
        fill=saltcolor,
        draw=white,
        line width=0.6pt
    }
] coordinates {
    (0.067778,44.4188) +- (0,0.9849)
};


\node[
    anchor=west,
    font=\scriptsize\bfseries,
    text=saltcolor,
    align=left,
    inner sep=1pt
] at (axis cs:0.13,88.8) {
    SALT\\[-1pt]
    {\normalfont 69--465$\times$ fewer parameters}
};


\addlegendimage{
    only marks,
    mark=square*,
    mark size=2.2pt,
    draw=fifacolor,
    fill=fifacolor
}
\addlegendentry{FIFA-RS}

\addlegendimage{
    only marks,
    mark=square*,
    mark size=2.2pt,
    draw=visualcolor,
    fill=visualcolor
}
\addlegendentry{VisualAD}

\addlegendimage{
    only marks,
    mark=square*,
    mark size=2.2pt,
    draw=avacolor,
    fill=avacolor
}
\addlegendentry{AVA-DINO}

\addlegendimage{
    only marks,
    mark=square*,
    mark size=2.2pt,
    draw=moecolor,
    fill=moecolor
}
\addlegendentry{MoECLIP}

\addlegendimage{
    only marks,
    mark=square*,
    mark size=2.2pt,
    draw=saltcolor,
    fill=saltcolor
}
\addlegendentry{SALT}

\end{axis}
\end{tikzpicture}

\caption{\textbf{Localization performance versus adaptation size.}
Pixel-level AP versus trainable parameters (logarithmic scale).
Colors identify the methods; circles and triangles denote
LEVIR-CD$\rightarrow$WHU and WHU$\rightarrow$LEVIR-CD,
respectively. Thin vertical lines connect each method's two
transfer results, and error bars show standard deviations
over three seeds. SALT achieves the highest pixel AP in
both directions with $69\times$ to $465\times$ fewer trainable
parameters than the reproduced baselines.}

\label{fig:performance_efficiency}
\end{figure}

Change detection in remote sensing typically relies on two spatially aligned images of the same area acquired at different times~\cite{peng2025deep,chen2020spatial}. In practical applications, however, a suitable pre-event image may be unavailable, outdated, or difficult to align with the current observation. Recent approaches, such as FIFA-RS~\cite{lin2026fifa} and BSSRS~\cite{lin2026bridging}, address this limitation by learning structural information from a labeled source dataset and transferring it to unseen target imagery using only the current image. Unlike conventional bi-temporal change detection, this setting does not directly determine whether a structure has changed over time. Instead, the objective is to localize structural evidence defined by the source annotations, such as changed buildings in LEVIR-CD~\cite{chen2020spatial} or building footprints in WHU~\cite{ji2018fully}.

The main challenge is to learn source-specific structural information while preserving the generalization ability of pretrained visual representations. Features extracted from vision foundation models provide rich visual information, but may not directly distinguish the positive structures defined by the source task from the background. Adapting these features using source annotations can improve this distinction, but excessive adaptation may also distort useful pretrained representations and reduce their transferability to unseen datasets. This trade-off becomes particularly important when the source and target datasets differ in appearance, acquisition conditions, or annotation semantics.

Existing approaches address this challenge through different adaptation strategies. FIFA-RS~\cite{lin2026fifa} adapts frozen CLIP~\cite{radford2021learning} representations for single-temporal structural anomaly reasoning, while BSSRS~\cite{lin2026bridging} combines CLIP and DINOv3~\cite{simeoni2025dinov3} features through constrained residual blending. Related visual anomaly detection methods also learn task-specific feature transformations. VisualAD~\cite{Hou_2026_CVPR} combines learned visual states with MLP-based feature recalibration, MoECLIP~\cite{Park_2026_CVPR} employs specialized low-rank experts, and AVA-DINO~\cite{11629903} uses separate adaptation branches. Although these methods offer different ways to adapt pretrained representations, the adaptation directions are generally learned independently of the states or representations used for prediction.

This separation motivates a more direct connection between feature adaptation and structural prediction. Instead of learning arbitrary correction directions, feature adaptation can be constrained by the same structural states used to distinguish positive and negative regions. Such a constraint limits how pretrained features can change while allowing task-specific adjustments. The central idea is therefore to determine not only how much each feature should be adapted, but also which directions are relevant to the structural prediction task.

To this end, this paper introduces SALT (State-Aligned Low-Rank Transfer), a framework that couples feature adaptation with structural prediction. SALT uses a frozen DINOv3 backbone and learns two structural states representing negative and positive evidence. These states serve two purposes: they are used to score each patch and to define the subspace in which its feature representation can be corrected. State-Aligned Low-Rank Adaptation (SALA), is introduced, to restrict feature corrections to this rank-2 subspace, while a lightweight predictor determines the correction coefficients for each patch. Directional State Alignment (DSA) guides these corrections to increase positive and decrease negative structural evidence, whereas Geometry Preservation limits changes to pairwise relationships among pretrained patch features. The main contributions of this paper are summarized as follows:

\begin{itemize}
    \item \textbf{State-Aligned Low-Rank Adaptation (SALA)} couples structural prediction and feature adaptation. The same learned positive and negative states used for prediction define the rank-2 subspace in which patch features are corrected, limiting adaptation to state-derived directions.

    \item \textbf{Directional State Alignment (DSA)} guides feature corrections toward increasing positive and decreasing negative structural evidence, while \textbf{Geometry Preservation} limits changes to pairwise relations among pretrained patch features.

    \item \textbf{Cross-dataset experiments} on LEVIR-CD and WHU demonstrate superior pixel-level localization over four reproduced baselines in both transfer directions under a common source-only protocol. SALT achieves the highest pixel-level AP with $69\times$--$465\times$ fewer trainable parameters (Fig.~\ref{fig:performance_efficiency}), while ablation studies support the benefits of state-aligned adaptation.
\end{itemize}

\section{Related Work}
\label{sec:related_work}

\noindent\textbf{Remote sensing change detection with reduced temporal supervision.}
Remote sensing change detection aims to identify regions that have changed between observations of the same area acquired at different times~\cite{peng2025deep,daudt2018fully}. Conventional approaches therefore typically rely on aligned bi-temporal imagery. In this context, several works have investigated how to reduce the need for paired supervision during training. ChangeStar~\cite{Zheng_2021_ICCV} learns from unpaired images with semantic annotations, while Self-Pair~\cite{Seo_2023_WACV} creates training pairs from a single source image. Changen~\cite{zheng2023scalable} and Changen2~\cite{zheng2024changen2} instead generate synthetic temporal changes and their corresponding labels. AnyChange~\cite{zheng2024segment} removes task-specific training through bitemporal latent matching in SAM, while SCL~\cite{du2026scl} combines single-temporal training data with vision-language prompt learning to generalize across change detection datasets without target-domain training. Foundation models have also been incorporated into conventional change detection, including ChangeCLIP~\cite{dong2024changeclip} and BAN~\cite{li2024new}. Nevertheless, these methods still require observations from two different times to produce the final change map.

\medskip
\noindent\textbf{Single-image remote sensing anomaly and structural localization.}
A separate line of research removes this temporal requirement and detects anomalous or structurally distinctive regions directly from a single remote sensing image. Li~\etal~\cite{li2024learning} learn a transferable deviation metric that detects anomalies across different remote sensing modalities in a zero-shot manner, while Xu~\etal~\cite{xu2024toward} combine large vision models with lightweight priors for on-orbit Earth surface anomaly detection. RSAD-CLIP~\cite{zhang2025rsad} further investigates vision-language representations for zero-shot Earth surface anomaly detection. More closely related to the considered setting, FIFA-RS~\cite{lin2026fifa} formulates change detection as single-temporal structural anomaly reasoning, using a frozen CLIP backbone, high-pass structural enhancement, and learnable multi-scale fusion. BSSRS~\cite{lin2026bridging} further addresses cross-dataset generalization by combining frozen CLIP and DINOv3 features, through constrained residual blending, to reduce structural over-specialization. FIFA-RS and BSSRS are the works closest to the considered protocol, in which only the current image is available and knowledge learned from a source dataset is directly transferred to an unseen target one. Both report F1 and IoU at the threshold that maximizes F1 on the target data, whereas the present study selects the decision threshold on the source validation split.

\medskip
\noindent\textbf{Zero-shot anomaly detection with vision-language models.}
Similar generalization challenges have been extensively investigated in zero-shot anomaly detection. Vision-language approaches typically represent normal and anomalous concepts through textual prompts and compare them with visual features. WinCLIP~\cite{jeong2023winclip} combines handcrafted state words and prompt templates with multi-scale CLIP features, while AnomalyCLIP~\cite{zhou2024anomalyclip} learns object-agnostic normal and anomalous prompts. AdaCLIP~\cite{cao2024adaclip} introduces static and image-dependent learnable prompts, whereas AA-CLIP~\cite{ma2025aa} learns anomaly-aware textual anchors together with visual adaptation. Despite their strong generalization ability, these approaches still depend on textual representations and cross-modal alignment. This has motivated more recent methods to model anomaly evidence directly in the visual feature space.

\medskip
\noindent\textbf{Visual anomaly modeling and feature adaptation.}
VisualAD~\cite{Hou_2026_CVPR} removes the language branch and learns visual states for normal and anomalous patterns together with an MLP-based feature recalibration. AnomalyVFM~\cite{Fucka_2026_CVPR} adapts vision foundation models using low-rank feature adapters, while MoECLIP~\cite{Park_2026_CVPR} performs patch-specific adaptation through dynamically routed LoRA~\cite{hu2022lora} experts. AULoRA~\cite{oh2025aulora} also introduces low-rank adaptation into the CLIP visual encoder, while AVA-DINO~\cite{11629903} employs separate normal and anomalous adaptation branches with text-guided routing. Overall, these methods demonstrate the benefits of lightweight visual adaptation for anomaly discrimination. However, their adaptation mechanisms are learned separately from the visual states or representations used for the final prediction. Moreover, previous studies have shown that strong adaptation of pretrained representations can reduce their robustness under distribution shift~\cite{kumar2022finetuning,Wortsman_2022_CVPR}. Parameter-efficient alternatives keep the backbone frozen, such as Rein~\cite{wei2024stronger}, which refines frozen features with learnable tokens for domain-generalized semantic segmentation, and LoReFT~\cite{wu2024reft}, which edits hidden representations within a learned low-rank subspace. SALT follows this representation-editing view, but instead of learning the subspace independently, it constructs the adaptation space directly from the learned structural states, so that the same states define both the prediction and the directions in which each feature can be corrected.

\section{Proposed Approach}
\label{sec:method}

\begin{figure*}[t]
  \centering
  \includegraphics[width=\textwidth]{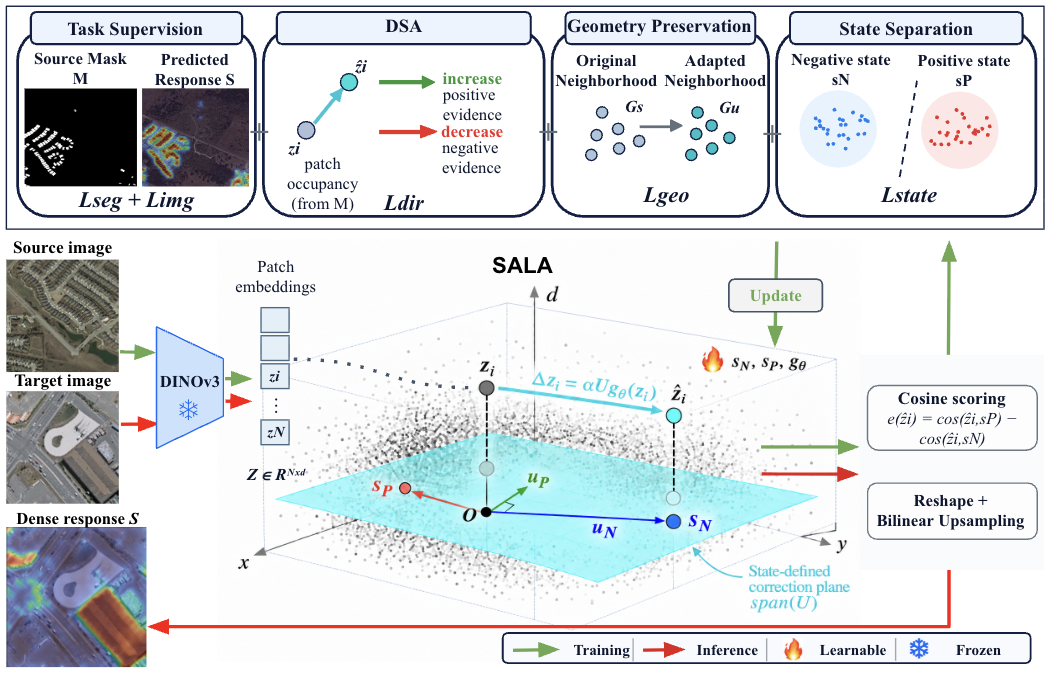}
  \caption{\textbf{Overview of SALT.}
A frozen DINOv3 extracts patch features from a single image.
SALA uses the learned negative and positive structural states
to define a rank-2 adaptation space, within which each patch
feature is corrected. The adapted features are then scored
to produce a dense structural response map.
The four training objectives (top) supervise structural
predictions, guide feature corrections, preserve feature
relationships, and separate the learned states.
Green arrows indicate training on the source dataset,
while red arrows indicate inference on the unseen target
dataset. Only the structural states and the lightweight
coefficient predictor are trained.}
  \label{fig:overview}
\end{figure*}

\subsection{Overview}
\label{sec:overview}

SALT learns task-specific structural information from a labeled source dataset, while limiting changes to the pretrained visual representation. Its overall architecture is illustrated in Fig.~\ref{fig:overview}. Given a single current image, a frozen DINOv3 backbone~\cite{simeoni2025dinov3} extracts dense patch features, and two learnable structural states represent negative and positive structural evidence in this feature space (Section~\ref{sec:feature_state}). State-Aligned Low-Rank Adaptation (SALA) corrects each patch feature within the rank-2 subspace spanned by the same states, with a lightweight coefficient predictor determining the magnitude of the correction along each direction (Section~\ref{sec:sala}). Directional State Alignment (DSA) supervises these corrections to raise the evidence of positive patches and lower that of negative ones (Section~\ref{sec:dsa}), while Geometry Preservation limits changes to the pairwise relations among the pretrained patch features (Section~\ref{sec:geometry}).

\subsection{Problem Formulation}
\label{sec:problem}

The task considered is single-temporal cross-dataset structural localization. Given a labeled source dataset containing images and binary segmentation masks, the objective is to learn structural representations that generalize to an unseen target domain. Image-level labels indicate the presence or absence of annotated positive regions, which are defined by the source task. At inference, only a single target image $\mathbf{x}^{t}$ is available. The model produces a dense structural response map $\mathbf{S}^{t}$ and an image-level score $r^{t}$, without access to a pre-event observation. All model parameters remain fixed, and binary predictions use a threshold selected on source validation data. No target-domain training, adaptation, or calibration is performed.

\subsection{Visual Feature Extraction and State Modeling}
\label{sec:feature_state}

Given an input image $\mathbf{x}$, a frozen DINOv3 backbone $\Phi$~\cite{simeoni2025dinov3} extracts patch features from its final transformer layer, excluding class and register tokens:
\begin{equation}
\mathbf{Z} = \Phi(\mathbf{x}) = [\mathbf{z}_1,\ldots,\mathbf{z}_N]^{\top} \in\mathbb{R}^{N\times d},
\label{eq:dino_features}
\end{equation}
where $\mathbf{z}_i\in\mathbb{R}^{d}$ denotes the feature vector of the $i$-th $16\times16$ image patch.

Following VisualAD~\cite{Hou_2026_CVPR}, two learnable structural states $\mathbf{s}_{\mathrm{N}},\mathbf{s}_{\mathrm{P}}\in\mathbb{R}^{d}$ represent negative and positive regions. Each patch is scored by comparing its cosine similarity with the two states:
\begin{equation}
e(\mathbf{z}) = \cos(\mathbf{z},\mathbf{s}_{\mathrm{P}}) - \cos(\mathbf{z},\mathbf{s}_{\mathrm{N}}),
\label{eq:structural_evidence}
\end{equation}
where higher scores indicate stronger positive structural evidence. Unlike VisualAD, which uses a separate MLP for feature adaptation, SALT also uses these states to define the adaptation space described next.

\subsection{State-Aligned Low-Rank Adaptation}
\label{sec:sala}

State-Aligned Low-Rank Adaptation (SALA) uses the learned structural states to guide feature adaptation. Since the negative and positive states are also used for prediction, they define the directions in which patch features can be modified. This restricts adaptation to a subspace of at most two dimensions, rather than allowing arbitrary changes in the original $d$-dimensional feature space. Unlike LoRA~\cite{hu2022lora}, which modifies model weights, SALA directly adapts patch features while keeping the backbone frozen.

Given the structural states $\mathbf{s}_{\mathrm{N}}$ and $\mathbf{s}_{\mathrm{P}}$, a basis is constructed using Gram--Schmidt orthogonalization:
\begin{equation}
\begin{aligned}
\mathbf{u}_{\mathrm{N}} &= \frac{\mathbf{s}_{\mathrm{N}}}{\|\mathbf{s}_{\mathrm{N}}\|_{2}+\epsilon},
\qquad
\widetilde{\mathbf{s}}_{\mathrm{P}} = \mathbf{s}_{\mathrm{P}}-(\mathbf{u}_{\mathrm{N}}^{\top}\mathbf{s}_{\mathrm{P}})\mathbf{u}_{\mathrm{N}},\\
\mathbf{u}_{\mathrm{P}} &= \frac{\widetilde{\mathbf{s}}_{\mathrm{P}}}{\|\widetilde{\mathbf{s}}_{\mathrm{P}}\|_{2}+\epsilon},
\qquad
\mathbf{U} = [\mathbf{u}_{\mathrm{N}},\mathbf{u}_{\mathrm{P}}]\in\mathbb{R}^{d\times2},
\end{aligned}
\label{eq:sala_basis}
\end{equation}
where $\epsilon$ ensures numerical stability. The two basis vectors span the state-defined subspace when the states are linearly independent (Supplementary Section~\ref{sec:supp_state_analysis}).

For each patch feature $\mathbf{z}_i$, a lightweight two-layer MLP $g_{\theta}$ predicts two coefficients $\mathbf{c}_i=g_{\theta}(\mathbf{z}_i)=[c_{i,\mathrm{N}},c_{i,\mathrm{P}}]^{\top}$. These coefficients determine the correction along each basis direction:
\begin{equation}
\widehat{\mathbf{z}}_{i}
= \mathbf{z}_{i} + \alpha\mathbf{U}\mathbf{c}_{i}
= \mathbf{z}_{i} + \alpha
\left(c_{i,\mathrm{N}}\mathbf{u}_{\mathrm{N}}
+ c_{i,\mathrm{P}}\mathbf{u}_{\mathrm{P}}\right),
\label{eq:sala}
\end{equation}
where $\alpha$ controls the correction magnitude.

The basis $\mathbf{U}$ is shared across patches, while the coefficients are predicted individually. Consequently, all feature corrections lie within the same state-defined subspace, and the correction matrix has rank at most two. This directly connects feature adaptation to the structural states used for prediction, rather than learning independent adaptation directions.

\subsection{Directional State Alignment}
\label{sec:dsa}

SALA determines the directions in which patch features can be adapted, but does not ensure that these corrections improve structural predictions. Directional State Alignment (DSA) addresses this by encouraging positive patches to gain structural evidence and negative patches to lose it. For each patch, the change in evidence is defined as:
\begin{equation}
\Delta e_i = e(\widehat{\mathbf{z}}_i) -
\operatorname{sg}\bigl(e(\mathbf{z}_i)\bigr),
\label{eq:evidence_change}
\end{equation}
where $\operatorname{sg}$ denotes stop-gradient. Although the backbone is frozen, the original evidence depends on the learnable structural states. Detaching this reference allows DSA to optimize the adapted evidence while updating the states and coefficient predictor through the adapted branch.

To account for patches containing both positive and negative pixels, the source mask $\mathbf{M}$ is average-pooled into patch-level positive occupancies $a_i=\operatorname{AvgPool}(\mathbf{M})_i\in[0,1]$, using a $16\times16$ kernel and stride. These occupancies weight the positive and negative directional losses:
\begin{equation}
\begin{aligned}
\mathcal{L}_{\mathrm{P}} &=
\frac{\sum_i a_i[m-\Delta e_i]_+}
{\sum_i a_i+\epsilon},\\
\mathcal{L}_{\mathrm{N}} &=
\frac{\sum_i(1-a_i)[m+\Delta e_i]_+}
{\sum_i(1-a_i)+\epsilon},\\
\mathcal{L}_{\mathrm{dir}} &=
\frac{1}{2}(\mathcal{L}_{\mathrm{P}}+\mathcal{L}_{\mathrm{N}}),
\end{aligned}
\label{eq:dsa}
\end{equation}
where $[x]_+=\max(0,x)$ and $m$ is the directional margin. The loss encourages $\Delta e_i\geq m$ for positive patches and $\Delta e_i\leq-m$ for negative patches, while mixed patches contribute to both terms according to their positive occupancy. The separate normalization reduces the effect of class imbalance. Unlike conventional deviation-based margin losses~\cite{pang2019deep}, which constrain the final anomaly score, DSA supervises the change in structural evidence caused by adaptation. Thus, SALA controls where features can change, while DSA guides the effect of these changes on structural predictions.

\subsection{Geometry Preservation}
\label{sec:geometry}

Although SALA restricts feature corrections to the state-defined subspace, these corrections can still change the relationships between patch features. Geometry Preservation limits such changes by encouraging pairwise feature similarities to remain consistent before and after adaptation. This follows the principle of similarity-preserving knowledge distillation~\cite{park2019relational,tung2019similarity}.

Let $\mathcal{R}$ denote a set of patch pairs sampled randomly from each image during training. Geometry Preservation minimizes the difference between their original and adapted cosine similarities:
\begin{equation}
\mathcal{L}_{\mathrm{geo}} = \frac{1}{|\mathcal{R}|} \sum_{(i,j)\in\mathcal{R}} \left| \cos(\widehat{\mathbf{z}}_{i},\widehat{\mathbf{z}}_{j}) - \cos(\mathbf{z}_{i},\mathbf{z}_{j}) \right|.
\label{eq:geometry}
\end{equation}
The loss is averaged over the images in each mini-batch. Unlike directly penalizing feature changes, this objective preserves their relative similarities while allowing task-specific corrections. It therefore balances DSA, which encourages changes in structural evidence, with their respective weights controlled by $\lambda_{\mathrm{dir}}$ and $\lambda_{\mathrm{geo}}$ in Eq.~\eqref{eq:total_loss}.

\subsection{Training and Inference}
\label{sec:training}

\noindent\textbf{Source training.}
The DINOv3 backbone remains frozen, while only the structural states $\mathbf{s}_{\mathrm{N}}$, $\mathbf{s}_{\mathrm{P}}$ and the coefficient predictor $g_{\theta}$ are trained on labeled source data ($0.068$M trainable parameters). The adapted patch features are scored using $e(\widehat{\mathbf{z}}_i)$, and their responses are reshaped and bilinearly upsampled to produce the dense map $\mathbf{S}$. Following VisualAD~\cite{Hou_2026_CVPR}, pixel-level predictions $\sigma(\mathbf{S})$ are supervised using focal~\cite{lin2017focal} and Dice~\cite{sudre2017generalised} losses, with $\mathcal{L}_{\mathrm{seg}}=\mathcal{L}_{\mathrm{focal}}+\mathcal{L}_{\mathrm{dice}}$. An image-level score $r$ is obtained by averaging the highest $\rho HW$ responses in $\mathbf{S}$ and supervised using binary cross-entropy, $\mathcal{L}_{\mathrm{img}}=\mathrm{BCE}(\sigma(r),y)$, where $\sigma$ denotes the sigmoid function.

A state-separation loss adopted from VisualAD~\cite{Hou_2026_CVPR}, $\mathcal{L}_{\mathrm{state}}=\max(0,\cos(\mathbf{s}_{\mathrm{P}},\mathbf{s}_{\mathrm{N}})+\tau)$, encourages the two structural states to remain distinct. The complete training objective combines these losses with DSA and Geometry Preservation:
\begin{equation}
\mathcal{L} = \mathcal{L}_{\mathrm{seg}} + \mathcal{L}_{\mathrm{img}} + \lambda_{\mathrm{state}}\mathcal{L}_{\mathrm{state}} + \lambda_{\mathrm{dir}}\mathcal{L}_{\mathrm{dir}} + \lambda_{\mathrm{geo}}\mathcal{L}_{\mathrm{geo}}.
\label{eq:total_loss}
\end{equation}

\noindent\textbf{Inference.}
At inference, all parameters remain fixed. The dense response map $\mathbf{S}$ is computed as during training and optionally smoothed using a Gaussian kernel with standard deviation $\sigma_g$. Pixel-level scores are taken directly from this map, without sigmoid transformation or normalization. Image-level scores use the same top-response aggregation, while binary predictions use a threshold selected on source validation data. No target-domain adaptation or calibration is performed.

\section{Experimental Results}
\label{sec:experiments}

\begin{table*}[t]
\centering
\caption{Cross-dataset localization performance (\%, mean$\pm$standard deviation over three seeds). All methods are reproduced under the common source-only protocol, and F1 and IoU use the threshold selected on the source validation split. Best results are in bold and second-best are underlined.}
\label{tab:comparison}

\footnotesize
\setlength{\tabcolsep}{2.8pt}
\renewcommand{\arraystretch}{0.96}

\begin{tabular}{@{}lcccccc@{}}
\toprule
\textbf{Method} &
\textbf{P-AUROC} &
\textbf{P-AP} &
\textbf{F1} &
\textbf{IoU} &
\textbf{I-AUROC} &
\textbf{I-AP} \\
\midrule

\multicolumn{7}{c}{\textbf{LEVIR-CD$\rightarrow$WHU}} \\
\midrule

FIFA-RS~\cite{lin2026fifa}
& $94.28\pm0.35$
& $68.76\pm1.56$
& $50.19\pm3.78$
& $33.56\pm3.33$
& $\mathbf{98.89\pm0.01}$
& $\mathbf{99.58\pm0.01}$ \\

VisualAD~\cite{Hou_2026_CVPR}
& $95.48\pm0.71$
& $74.28\pm1.90$
& $55.15\pm1.97$
& $38.09\pm1.89$
& $\underline{98.81\pm0.43}$
& $\underline{99.56\pm0.15}$ \\

AVA-DINO~\cite{11629903}
& $\underline{96.91\pm0.20}$
& $\underline{78.32\pm0.09}$
& $\underline{55.26\pm7.23}$
& $\underline{38.41\pm6.83}$
& $98.46\pm0.20$
& $99.45\pm0.07$ \\

MoECLIP~\cite{Park_2026_CVPR}
& $93.27\pm0.16$
& $60.48\pm0.72$
& $42.91\pm0.88$
& $27.32\pm0.72$
& $98.68\pm0.07$
& $99.53\pm0.02$ \\

SALT
& $\mathbf{98.08\pm0.08}$
& $\mathbf{87.77\pm0.21}$
& $\mathbf{79.59\pm0.31}$
& $\mathbf{66.10\pm0.42}$
& $98.34\pm0.34$
& $99.43\pm0.12$ \\

\midrule
\multicolumn{7}{c}{\textbf{WHU$\rightarrow$LEVIR-CD}} \\
\midrule

FIFA-RS~\cite{lin2026fifa}
& $88.14\pm0.78$
& $28.67\pm1.51$
& $13.84\pm0.47$
& $7.44\pm0.27$
& $\mathbf{91.61\pm0.69}$
& $\mathbf{99.15\pm0.08}$ \\

VisualAD~\cite{Hou_2026_CVPR}
& $92.73\pm0.28$
& $37.29\pm0.43$
& $\underline{27.35\pm1.12}$
& $\underline{15.84\pm0.75}$
& $78.33\pm4.50$
& $97.62\pm0.69$ \\

AVA-DINO~\cite{11629903}
& $\underline{93.67\pm0.08}$
& $\underline{38.63\pm0.28}$
& $21.92\pm1.67$
& $12.32\pm1.05$
& $84.80\pm2.95$
& $98.37\pm0.41$ \\

MoECLIP~\cite{Park_2026_CVPR}
& $88.87\pm0.23$
& $27.14\pm0.65$
& $12.67\pm0.21$
& $6.76\pm0.12$
& $89.89\pm0.88$
& $98.88\pm0.14$ \\

SALT
& $\mathbf{94.82\pm0.18}$
& $\mathbf{44.42\pm0.98}$
& $\mathbf{46.24\pm2.19}$
& $\mathbf{30.09\pm1.84}$
& $\underline{90.27\pm1.61}$
& $\underline{98.94\pm0.11}$ \\

\bottomrule
\end{tabular}
\end{table*}

\subsection{Experimental Setup}
\label{sec:setup}

\noindent\textbf{Datasets.} Experiments are conducted on LEVIR-CD~\cite{chen2020spatial} and the aerial subset of the WHU Building Dataset~\cite{ji2018fully} (hereafter WHU), following the single-temporal setting of FIFA-RS~\cite{lin2026fifa} and BSSRS~\cite{lin2026bridging}. LEVIR-CD contains 637 aligned bi-temporal image pairs of size $1024\times1024$ at $0.5$ m/pixel with binary building-change annotations, split into 445 training, 64 validation, and 128 test pairs. Only the post-event image of each pair is used, with the change mask as the positive annotation. WHU contains 8,189 tiles of size $512\times512$ at $0.3$ m/pixel with building-footprint masks, split into 4,736 training, 1,036 validation, and 2,416 test tiles. The two datasets therefore differ in positive semantics, since LEVIR-CD annotates changed buildings whereas WHU annotates all buildings in the current image, and also in ground sampling distance, since resizing to the network input yields approximately $1.0$ and $0.3$ m/pixel, respectively. The evaluation thus measures cross-dataset transfer of source-defined structural evidence, rather than conventional temporal change detection.

\noindent\textbf{Experimental setup and evaluation protocol.}
SALT uses a frozen DINOv3 ViT-L/16 backbone with an input
resolution of $518\times518$. Evaluation is conducted in both
directions, LEVIR-CD$\rightarrow$WHU and WHU$\rightarrow$LEVIR-CD.
For each direction, hyperparameters are selected on the source
validation split using seed 111, and the resulting configuration
is evaluated over three seeds (111, 222, and 333).
Training, checkpoint selection, and threshold estimation use only
source data, without target-domain adaptation or calibration.
Performance is measured using pixel-level AUROC (P-AUROC),
average precision (P-AP), F1, and IoU, alongside image-level
AUROC (I-AUROC) and AP (I-AP).
Unlike FIFA-RS and BSSRS, F1 and IoU use thresholds selected
by maximizing F1 on source validation rather than target data.
Further implementation details and hyperparameter analyses
are provided in Supplementary Section~\ref{sec:supp_configuration}.

\subsection{Comparative Evaluation}
\label{sec:comparison}

%
%

\begin{figure*}[t]
    \centering
    \includegraphics[width=\textwidth]
    {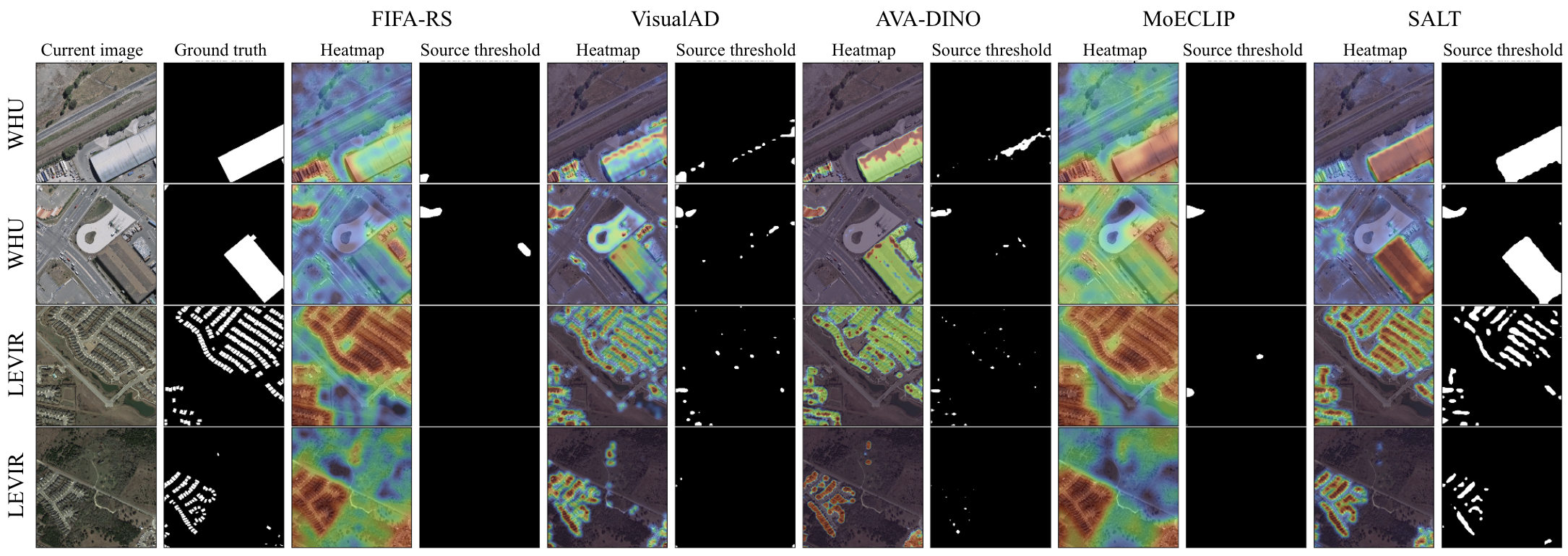}
     \caption{Indicative cross-dataset localization results. Rows are labeled by the target dataset (WHU rows from LEVIR-CD$\rightarrow$WHU, LEVIR rows from WHU$\rightarrow$LEVIR-CD, with LEVIR denoting LEVIR-CD). For each method, the heatmap shows the dense response map and the binary mask uses the threshold selected on the source validation split.}
    \label{fig:qualitative_comparison}
\end{figure*}

\noindent\textbf{Baseline methods and reported performance.}
SALT is compared with four recent methods, namely FIFA-RS~\cite{lin2026fifa}, VisualAD~\cite{Hou_2026_CVPR}, AVA-DINO~\cite{11629903}, and MoECLIP~\cite{Park_2026_CVPR}, all reproduced under the common source-only protocol of Section~\ref{sec:setup} with identical splits and seeds. Baselines use their official implementations and author-recommended hyperparameters, whereas SALT hyperparameters are selected on source validation. All methods use source-selected decision thresholds. VisualAD, originally built on CLIP and DINOv2 ViT-L/14 backbones, and AVA-DINO use the same frozen DINOv3 ViT-L/16 backbone as SALT, whereas FIFA-RS and MoECLIP retain their original CLIP encoders, and methods requiring text use the remote sensing prompt set of FIFA-RS. The reproduced FIFA-RS reaches P-AUROC/P-AP of $94.28/68.76$ and $88.14/28.67$ in LEVIR-CD$\rightarrow$WHU and WHU$\rightarrow$LEVIR-CD, close to the $94.91/70.36$ and $88.63/28.06$ reported in~\cite{lin2026fifa}, while its lower F1 reflects the source-selected threshold. BSSRS~\cite{lin2026bridging} could not be reproduced, since its linked repository contained no code at the time of writing. For reference, it reports threshold-independent P-AUROC/P-AP of $95.38/71.00$ and $90.41/31.74$ for the two directions, obtained with its own training setup,  and SALT exceeds both values in each direction ($98.08/87.77$ and $94.82/44.42$), although the evaluation setups are not identical.

\noindent\textbf{Discussion.}
From Table~\ref{tab:comparison}, Table~\ref{tab:computational_cost},
and Fig.~\ref{fig:qualitative_comparison}, the following observations
can be made:
a) SALT achieves the highest pixel-level performance in both transfer directions. Compared with the strongest reproduced baselines, P-AP improves by $9.45$ and $5.79$ points, F1 by $24.33$ and $18.89$ points, and IoU by $27.69$ and $14.25$ points for LEVIR-CD$\rightarrow$WHU and WHU$\rightarrow$LEVIR-CD, respectively.
b) VisualAD and AVA-DINO use the same frozen DINOv3 backbone as SALT, suggesting that the improvements cannot be attributed solely to
the backbone.
c) Image-level performance is comparable across methods, with I-AP exceeding $97.6\%$ in all cases. Notably, $71.65\%$ of WHU and $92.19\%$ of LEVIR-CD test images contain positive structures, contributing to the high I-AP values. Although SALT does not achieve the highest image-level scores, its I-AUROC remains within $0.55$
and $1.34$ points of the best results in the two directions,
respectively, while the corresponding I-AP differences are only $0.15$ and $0.21$ points. d) WHU$\rightarrow$LEVIR-CD remains the more challenging transfer
direction, consistent with the different annotation semantics of the two datasets. e) SALT requires only $0.068$M trainable parameters, approximately
$70$ times fewer than the lightest baseline, while achieving the lowest inference latency and peak memory usage among the compared implementations.
f) As illustrated in Fig.~\ref{fig:qualitative_comparison}, SALT produces more complete localization maps in the selected examples, where competing methods often recover only parts of the relevant structures under source-selected thresholds.

\begin{table}[t]
\centering
\caption{\textbf{Computational cost comparison.}
Trainable parameters, non-backbone GFLOPs, training time,
peak inference VRAM, and latency.
Training time is reported for LEVIR-CD$\rightarrow$WHU
(L$\rightarrow$W, minutes) and WHU$\rightarrow$LEVIR-CD
(W$\rightarrow$L, hours). Memory and latency are measured
on an NVIDIA RTX 6000 Ada GPU.
}
\label{tab:computational_cost}

\centering
\setlength{\tabcolsep}{3pt}
\renewcommand{\arraystretch}{1.12}

\resizebox{\columnwidth}{!}{%
\begin{tabular}{@{}lcccccc@{}}
\toprule
\textbf{Method} &
\shortstack{\textbf{Params.}\\\textbf{(M)} $\downarrow$} &
\shortstack{\textbf{GFLOPs}\\$\downarrow$} &
\shortstack{\textbf{Train L$\rightarrow$W}\\\textbf{(min)} $\downarrow$} &
\shortstack{\textbf{Train W$\rightarrow$L}\\\textbf{(h)} $\downarrow$} &
\shortstack{\textbf{VRAM}\\\textbf{(GiB)} $\downarrow$} &
\shortstack{\textbf{Latency}\\\textbf{(ms)} $\downarrow$} \\
\midrule
FIFA-RS
& 20.665 & 45.361
& $29.1\pm0.6$ & $6.77\pm1.00$
& 2.21 & $132.4\pm0.4$ \\

VisualAD
& 31.521 & 51.758
& $16.0\pm2.9$ & $2.78\pm0.08$
& $\underline{1.69}$ & $\underline{43.4\pm1.3}$ \\

AVA-DINO
& 28.312 & $\underline{25.841}$
& $\underline{5.3\pm0.1}$ & $\underline{1.17\pm0.03}$
& 1.81 & $45.5\pm11.0$ \\

MoECLIP
& $\underline{4.675}$ & 28.456
& $31.8\pm1.4$ & $6.29\pm0.05$
& 2.00 & $76.4\pm2.4$ \\

\textbf{SALT}
& $\mathbf{0.068}$ & $\mathbf{0.139}$
& $\mathbf{5.1\pm0.3}$ & $\mathbf{1.09\pm0.03}$
& $\mathbf{1.22}$ & $\mathbf{40.2\pm1.2}$ \\
\bottomrule
\end{tabular}%
}
\end{table}

\subsection{Ablation Study}
\label{sec:ablation}

\noindent\textbf{Component ablation.}
To assess the contribution of each SALT component,
Table~\ref{tab:component_ablation} evaluates:
a) \textbf{Linear Probe}, which directly predicts
structural evidence from frozen DINOv3 patch features;
b) \textbf{Structural States}, which scores frozen
features using learned negative and positive states
without adaptation;
c) \textbf{SALA}, which additionally adapts features
within the state-defined rank-2 subspace; and
d) \textbf{SALA with DSA and Geometry Preservation},
evaluating each objective independently and jointly.
All configurations share the frozen backbone,
training settings, and source-only evaluation protocol,
with the applicable objectives used in each case.

\begin{table}[t]
\centering
\caption{Component ablation (\%, mean$\pm$standard
deviation over three seeds). Linear Probe and
Structural States are reference predictors.
The remaining rows evaluate combinations of
DSA and Geometry Preservation with SALA.
Best results are in bold.}
\label{tab:component_ablation}

\scriptsize
\setlength{\tabcolsep}{2pt}
\renewcommand{\arraystretch}{1.03}

\resizebox{\columnwidth}{!}{%
\begin{tabular}{@{}lcccccc@{}}
\toprule
\textbf{Configuration} &
\textbf{DSA} &
\textbf{Geo.} &
\textbf{P-AUROC} &
\textbf{P-AP} &
\textbf{F1} &
\textbf{IoU} \\
\midrule

\multicolumn{7}{c}{\textbf{LEVIR-CD$\rightarrow$WHU}} \\
\midrule

DINOv3 + Linear Probe
& -- & --
& $95.1\pm0.1$
& $76.3\pm0.6$
& $57.3\pm1.7$
& $40.2\pm1.6$ \\

Structural States
& -- & --
& $71.9\pm14.7$
& $16.4\pm4.0$
& $32.4\pm14.8$
& $19.9\pm11.2$ \\

\midrule

SALA
& No & No
& $96.6\pm0.1$
& $85.2\pm0.4$
& $76.5\pm0.3$
& $62.0\pm0.4$ \\

SALA + DSA
& Yes & No
& $97.0\pm0.0$
& $85.9\pm0.2$
& $79.3\pm0.2$
& $65.7\pm0.2$ \\

SALA + Geometry
& No & Yes
& $97.6\pm0.1$
& $86.4\pm0.3$
& $78.7\pm0.2$
& $64.9\pm0.3$ \\

SALT
& Yes & Yes
& $\mathbf{98.1\pm0.1}$
& $\mathbf{87.8\pm0.2}$
& $\mathbf{79.6\pm0.3}$
& $\mathbf{66.1\pm0.4}$ \\

\midrule
\multicolumn{7}{c}{\textbf{WHU$\rightarrow$LEVIR-CD}} \\
\midrule

DINOv3 + Linear Probe
& -- & --
& $89.3\pm0.2$
& $31.2\pm0.2$
& $37.3\pm0.2$
& $22.9\pm0.2$ \\

Structural States
& -- & --
& $84.7\pm0.0$
& $39.0\pm0.1$
& $31.8\pm1.0$
& $19.0\pm0.8$ \\

\midrule

SALA
& No & No
& $94.1\pm0.2$
& $41.8\pm0.8$
& $45.7\pm0.7$
& $29.6\pm0.6$ \\

SALA + DSA
& Yes & No
& $\mathbf{94.8\pm0.1}$
& $\mathbf{44.4\pm1.0}$
& $\mathbf{48.0\pm0.8}$
& $\mathbf{31.6\pm0.7}$ \\

SALA + Geometry
& No & Yes
& $94.5\pm0.3$
& $44.0\pm1.6$
& $46.5\pm2.4$
& $30.3\pm2.0$ \\

SALT
& Yes & Yes
& $\mathbf{94.8\pm0.2}$
& $\mathbf{44.4\pm1.0}$
& $46.2\pm2.2$
& $30.1\pm1.8$ \\

\bottomrule
\end{tabular}%
}
\end{table}

\medskip
\noindent\textbf{Discussion.}
Table~\ref{tab:component_ablation} supports three observations.
a) \textbf{Pretrained DINOv3 features provide a strong
baseline, but structural states alone are insufficient.}
The linear probe demonstrates the discriminative capability
of frozen DINOv3 features, while direct Structural States
perform substantially worse on LEVIR-CD$\rightarrow$WHU
and yield lower F1 and IoU in the reverse direction.
b) \textbf{SALA substantially improves localization.}
Adapting features within the state-defined subspace improves
all pixel-level metrics over both frozen-feature predictors,
increasing P-AP over the linear probe by $8.9$ and $10.6$
points in the two directions, respectively.
c) \textbf{DSA and Geometry Preservation further improve SALA.}
Both objectives independently improve all pixel-level
metrics in both directions. Their combination achieves the
best results on LEVIR-CD$\rightarrow$WHU, while DSA
alone yields higher F1 and IoU on WHU$\rightarrow$LEVIR-CD.
Further analyses of SALA and the learned state space,
DSA, and Geometry Preservation are provided in
Supplementary Sections~\ref{sec:supp_state_analysis}, \ref{sec:supp_dsa}, and \ref{sec:supp_geometry}, respectively.

\section{Limitations and Future Work}
\label{sec:limitations}
The current formulation represents structural evidence with a single pair of negative and positive states, which restricts SALA to a rank-2 adaptation space and may not capture heterogeneous positive appearances. Without a pre-event observation, the single-temporal setting cannot distinguish true temporal change from source-defined structural evidence, so that a model trained on building footprints also responds to buildings that predate the change, which likely contributes to the lower WHU$\rightarrow$LEVIR-CD performance. The evaluation is further limited to one dataset pair, two transfer directions, and three seeds, with source and target differing jointly in annotation semantics and ground sampling distance. Future work could investigate multiple context-dependent states, adaptive-rank adaptation, and broader transfer across semantic, geographic, and sensor shifts, including targets whose annotations match the source semantics, such as the WHU building change detection dataset~\cite{ji2018fully}.

\section{Conclusion}
\label{sec:conclusion}

This paper introduced SALT for single-temporal cross-dataset structural localization in remote sensing. SALT learns negative and positive structural states on a frozen DINOv3 backbone and uses the same states both to score each patch and to span the rank-2 subspace in which patch features are corrected. Directional State Alignment supervises how these corrections change the structural evidence, while Geometry Preservation limits changes to the pairwise relations among pretrained patch features. On LEVIR-CD and WHU, SALT achieves the highest pixel-level performance among four reproduced baselines in both transfer directions with only $0.068$M trainable parameters, and ablations show that state-aligned adaptation outperforms both unrestricted and generic low-rank adaptation. These results support defining feature adaptation through the same states used for prediction.

\section*{Acknowledgments}

This work was supported by the European Union's Horizon Europe research and innovation programme under Grant Agreement No. 101168042 (TRIFFID: auTonomous Robotic aId For increasing FIrst responDers efficiency).

{
    \small
    \bibliographystyle{ieeenat_fullname}
    \bibliography{bibliography}
}

\clearpage
\setcounter{section}{0}
\setcounter{subsection}{0}
\setcounter{figure}{0}
\setcounter{table}{0}
\setcounter{equation}{0}
\renewcommand{\thesection}{S\arabic{section}}
\renewcommand{\thesubsection}{S\arabic{section}.\arabic{subsection}}
\renewcommand{\thefigure}{S\arabic{figure}}
\renewcommand{\thetable}{S\arabic{table}}
\renewcommand{\theequation}{S\arabic{equation}}
\renewcommand{\theHsection}{supp.section.\arabic{section}}
\renewcommand{\theHsubsection}{supp.subsection.\arabic{section}.\arabic{subsection}}
\renewcommand{\theHfigure}{supp.figure.\arabic{figure}}
\renewcommand{\theHtable}{supp.table.\arabic{table}}
\renewcommand{\theHequation}{supp.equation.\arabic{equation}}
\maketitlesupplementary
\makeatletter\setlength{\@fptop}{0pt}\makeatother

\section{Supplementary Overview}
\label{sec:supp_overview}

This supplementary material provides additional
implementation details and experimental analyses
supporting the main paper.
Section~\ref{sec:supp_configuration} presents the
hyperparameter selection procedure and final model
configuration. Section~\ref{sec:supp_state_analysis}
compares different adaptation designs and examines
the learned structural states and feature corrections.
Sections~\ref{sec:supp_dsa} and~\ref{sec:supp_geometry}
further analyze Directional State Alignment and
Geometry Preservation, respectively.

\section{Model Configuration and Hyperparameter Analysis}
\label{sec:supp_configuration}

\noindent\textbf{Selection protocol.}
Hyperparameters are selected independently on the LEVIR-CD and WHU
source-validation splits using seed 111. Both datasets follow the
same staged procedure, where one parameter is varied at a time
while previously selected settings are retained. Selection is
guided primarily by P-AP, with P-AUROC also considered.
The following analyses report the individual stages, followed
by the final configurations used for the three-seed evaluation.
No target-domain data are used for hyperparameter selection
or calibration.

\medskip
\noindent\textbf{Backbone selection.}
The first stage compares CLIP ViT-L/14 and three DINOv3
backbones using their final-layer features. As reported in
Table~\ref{tab:supp_backbone}, DINOv3 ViT-L/16 achieves the
highest P-AP and P-AUROC on both datasets and is selected
for the remaining experiments. The improvement over
ViT-B/16 is relatively small on WHU.

\begin{table}[h]
\centering
\caption{Backbone comparison on source validation (\%).}
\label{tab:supp_backbone}
\footnotesize
\setlength{\tabcolsep}{3pt}
\renewcommand{\arraystretch}{1.05}
\begin{tabular}{@{}lcccc@{}}
\toprule
& \multicolumn{2}{c}{\textbf{LEVIR-CD}} &
  \multicolumn{2}{c}{\textbf{WHU}} \\
\cmidrule(lr){2-3}\cmidrule(lr){4-5}
\textbf{Backbone} &
\textbf{P-AUROC} & \textbf{P-AP} &
\textbf{P-AUROC} & \textbf{P-AP} \\
\midrule
CLIP ViT-L/14
& $93.69$ & $41.96$ & $97.43$ & $91.95$ \\
DINOv3 ViT-S/16
& $94.53$ & $40.06$ & $98.68$ & $92.98$ \\
DINOv3 ViT-B/16
& $95.39$ & $46.03$ & $99.02$ & $93.26$ \\
\textbf{DINOv3 ViT-L/16}
& $\mathbf{95.51}$ & $\mathbf{46.52}$ &
  $\mathbf{99.07}$ & $\mathbf{93.30}$ \\
\bottomrule
\end{tabular}
\end{table}

\medskip
\noindent\textbf{DSA loss weight.}
The weight $\lambda_{\mathrm{dir}}$ controls the contribution
of directional supervision. Table~\ref{tab:supp_lambda_dir}
shows that $\lambda_{\mathrm{dir}}=0.50$ gives the highest
P-AP on LEVIR-CD, while WHU favors $1.00$.
The effect is more pronounced on LEVIR-CD, whereas
WHU exhibits minor differences across the tested values.

\begin{table}[h]
\centering
\caption{Effect of the DSA loss weight on source validation (\%).}
\label{tab:supp_lambda_dir}
\footnotesize
\setlength{\tabcolsep}{3pt}
\renewcommand{\arraystretch}{1.05}
\begin{tabular}{@{}ccccc@{}}
\toprule
& \multicolumn{2}{c}{\textbf{LEVIR-CD}} &
  \multicolumn{2}{c}{\textbf{WHU}} \\
\cmidrule(lr){2-3}\cmidrule(lr){4-5}
$\boldsymbol{\lambda_{\mathrm{dir}}}$ &
\textbf{P-AUROC} & \textbf{P-AP} &
\textbf{P-AUROC} & \textbf{P-AP} \\
\midrule
$0.00$ & $94.78$ & $43.25$ & $99.07$ & $93.25$ \\
$0.10$ & $95.11$ & $44.21$ & $99.07$ & $93.28$ \\
$0.25$ & $95.18$ & $45.13$ & $99.07$ & $93.27$ \\
$0.50$ & $95.37$ & $\mathbf{45.65}$ &
         $99.07$ & $93.30$ \\
$1.00$ & $\mathbf{95.44}$ & $44.66$ &
         $\mathbf{99.08}$ & $\mathbf{93.35}$ \\
\bottomrule
\end{tabular}
\end{table}

\medskip
\noindent\textbf{DSA margin.}
After selecting the DSA weight, the directional
margin $m$ is examined. On LEVIR-CD, $m=0.10$
and $0.20$ yield nearly identical performance,
with the smaller margin retained for the final
configuration. On WHU, performance improves
with increasing margins, reaching its highest
P-AP at $m=0.20$ (Table~\ref{tab:supp_margin}).

\begin{table}[h]
\centering
\caption{Effect of the DSA margin on source validation (\%).}
\label{tab:supp_margin}
\footnotesize
\setlength{\tabcolsep}{3pt}
\renewcommand{\arraystretch}{1.05}
\begin{tabular}{@{}ccccc@{}}
\toprule
& \multicolumn{2}{c}{\textbf{LEVIR-CD}} &
  \multicolumn{2}{c}{\textbf{WHU}} \\
\cmidrule(lr){2-3}\cmidrule(lr){4-5}
$\boldsymbol{m}$ &
\textbf{P-AUROC} & \textbf{P-AP} &
\textbf{P-AUROC} & \textbf{P-AP} \\
\midrule
$0.000$ & $95.30$ & $44.92$ & $99.05$ & $93.20$ \\
$0.025$ & $95.37$ & $45.45$ & $99.07$ & $93.31$ \\
$0.050$ & $95.37$ & $45.65$ & $99.08$ & $93.35$ \\
$0.100$ & $95.40$ & $\mathbf{46.01}$ &
          $99.10$ & $93.47$ \\
$0.200$ & $\mathbf{95.41}$ & $\mathbf{46.01}$ &
          $\mathbf{99.11}$ & $\mathbf{93.57}$ \\
\bottomrule
\end{tabular}
\end{table}

\medskip
\noindent\textbf{Geometry Preservation weight.}
The regularization weight $\lambda_{\mathrm{geo}}$ is examined
after fixing the DSA settings. Both datasets achieve their
highest P-AP at $\lambda_{\mathrm{geo}}=0.10$
(Table~\ref{tab:supp_lambda_geo}). Larger values reduce P-AP,
suggesting that excessive regularization can restrict
task-specific feature corrections.

\begin{table}[h]
\centering
\caption{Effect of the Geometry Preservation weight on source validation (\%).}
\label{tab:supp_lambda_geo}
\footnotesize
\setlength{\tabcolsep}{3pt}
\renewcommand{\arraystretch}{1.05}
\begin{tabular}{@{}ccccc@{}}
\toprule
& \multicolumn{2}{c}{\textbf{LEVIR-CD}} &
  \multicolumn{2}{c}{\textbf{WHU}} \\
\cmidrule(lr){2-3}\cmidrule(lr){4-5}
$\boldsymbol{\lambda_{\mathrm{geo}}}$ &
\textbf{P-AUROC} & \textbf{P-AP} &
\textbf{P-AUROC} & \textbf{P-AP} \\
\midrule
$0.00$ & $95.35$ & $45.24$ & $99.07$ & $93.28$ \\
$0.05$ & $95.38$ & $45.68$ & $99.10$ & $93.49$ \\
$0.10$ & $95.40$ & $\mathbf{46.01}$ &
         $\mathbf{99.11}$ & $\mathbf{93.57}$ \\
$0.25$ & $\mathbf{95.45}$ & $44.76$ &
         $99.09$ & $93.36$ \\
$0.50$ & $95.35$ & $44.36$ & $99.01$ & $93.03$ \\
\bottomrule
\end{tabular}
\end{table}

\medskip
\noindent\textbf{Residual scaling factor.}
The residual scale $\alpha$ determines the magnitude of the
SALA correction. As shown in Table~\ref{tab:supp_alpha},
LEVIR-CD achieves its highest P-AP at $\alpha=0.25$.
On WHU, larger scales improve performance, with
$\alpha=2.00$ giving the highest P-AP, although its
advantage over $1.50$ is marginal.

\begin{table}[h]
\centering
\caption{Effect of the residual scaling factor on source validation (\%).}
\label{tab:supp_alpha}
\footnotesize
\setlength{\tabcolsep}{3pt}
\renewcommand{\arraystretch}{1.05}
\begin{tabular}{@{}ccccc@{}}
\toprule
& \multicolumn{2}{c}{\textbf{LEVIR-CD}} &
  \multicolumn{2}{c}{\textbf{WHU}} \\
\cmidrule(lr){2-3}\cmidrule(lr){4-5}
$\boldsymbol{\alpha}$ &
\textbf{P-AUROC} & \textbf{P-AP} &
\textbf{P-AUROC} & \textbf{P-AP} \\
\midrule
$0.25$ & $\mathbf{95.42}$ & $\mathbf{46.14}$ &
         $99.06$ & $93.27$ \\
$0.50$ & $95.36$ & $45.76$ & $99.07$ & $93.37$ \\
$1.00$ & $95.40$ & $46.01$ & $99.11$ & $93.57$ \\
$1.50$ & $95.36$ & $45.54$ &
         $\mathbf{99.12}$ & $93.57$ \\
$2.00$ & $95.37$ & $45.72$ &
         $\mathbf{99.12}$ & $\mathbf{93.58}$ \\
\bottomrule
\end{tabular}
\end{table}

\medskip
\noindent\textbf{Top-pooling ratio.}
The top-pooling ratio $\rho$ controls the fraction of
the strongest responses used for image-level aggregation.
The smallest evaluated ratio, $\rho=0.0025$, gives the
highest P-AP on both datasets (Table~\ref{tab:supp_topk}).
Increasing the ratio consistently reduces P-AP,
indicating that aggregation benefits from focusing
on the strongest patch responses.

\begin{table}[h]
\centering
\caption{Effect of the top-pooling ratio on source validation (\%).}
\label{tab:supp_topk}
\footnotesize
\setlength{\tabcolsep}{3pt}
\renewcommand{\arraystretch}{1.05}
\begin{tabular}{@{}ccccc@{}}
\toprule
& \multicolumn{2}{c}{\textbf{LEVIR-CD}} &
  \multicolumn{2}{c}{\textbf{WHU}} \\
\cmidrule(lr){2-3}\cmidrule(lr){4-5}
$\boldsymbol{\rho}$ &
\textbf{P-AUROC} & \textbf{P-AP} &
\textbf{P-AUROC} & \textbf{P-AP} \\
\midrule
$0.0025$ & $\mathbf{95.69}$ & $\mathbf{48.34}$ &
           $99.22$ & $\mathbf{94.60}$ \\
$0.0050$ & $95.61$ & $47.87$ &
           $\mathbf{99.23}$ & $94.58$ \\
$0.0075$ & $95.58$ & $47.41$ & $99.19$ & $94.22$ \\
$0.0100$ & $95.54$ & $47.21$ & $99.17$ & $94.05$ \\
$0.0150$ & $95.48$ & $46.67$ & $99.15$ & $93.79$ \\
$0.0200$ & $95.42$ & $46.14$ & $99.12$ & $93.58$ \\
\bottomrule
\end{tabular}
\end{table}

\medskip
\noindent\textbf{Gaussian smoothing.}
Finally, Gaussian smoothing is evaluated on the response maps.
On LEVIR-CD, moderate smoothing ($\sigma_{\mathrm{g}}=4$)
achieves the highest P-AP, whereas WHU performs best
without smoothing ($\sigma_{\mathrm{g}}=0$).
Stronger smoothing reduces P-AP on both datasets
(Table~\ref{tab:supp_sigma}).

\begin{table}[h]
\centering
\caption{Effect of Gaussian smoothing on source validation (\%).}
\label{tab:supp_sigma}
\footnotesize
\setlength{\tabcolsep}{3pt}
\renewcommand{\arraystretch}{1.05}
\begin{tabular}{@{}ccccc@{}}
\toprule
& \multicolumn{2}{c}{\textbf{LEVIR-CD}} &
  \multicolumn{2}{c}{\textbf{WHU}} \\
\cmidrule(lr){2-3}\cmidrule(lr){4-5}
$\boldsymbol{\sigma_{\mathrm{g}}}$ &
\textbf{P-AUROC} & \textbf{P-AP} &
\textbf{P-AUROC} & \textbf{P-AP} \\
\midrule
$0$ & $95.69$ & $48.14$ &
      $\mathbf{99.25}$ & $\mathbf{94.72}$ \\
$2$ & $\mathbf{95.70}$ & $48.27$ &
      $99.23$ & $94.63$ \\
$4$ & $95.69$ & $\mathbf{48.34}$ &
      $99.21$ & $94.56$ \\
$8$ & $95.43$ & $46.37$ &
      $99.07$ & $93.53$ \\
\bottomrule
\end{tabular}
\end{table}

\medskip
\noindent\textbf{Final configuration.}
The preceding analysis yields the source-specific settings
summarized in Table~\ref{tab:supp_final_config}.
Both datasets select the same backbone, Geometry Preservation
weight, and top-pooling ratio, while different DSA settings,
residual scales, and smoothing strengths are retained.
Each configuration is fixed for the final three-seed
cross-dataset evaluation.

\begin{table}[h]
\centering
\caption{Final source-selected SALT configurations.}
\label{tab:supp_final_config}
\footnotesize
\setlength{\tabcolsep}{5pt}
\renewcommand{\arraystretch}{1.08}
\begin{tabular}{@{}lcc@{}}
\toprule
\textbf{Parameter} &
\textbf{LEVIR-CD} &
\textbf{WHU} \\
\midrule
Backbone
& \multicolumn{2}{c}{DINOv3 ViT-L/16} \\
DSA weight $\lambda_{\mathrm{dir}}$
& $0.50$ & $1.00$ \\
DSA margin $m$
& $0.10$ & $0.20$ \\
Geometry weight $\lambda_{\mathrm{geo}}$
& $0.10$ & $0.10$ \\
Residual scale $\alpha$
& $0.25$ & $2.00$ \\
Top-pooling ratio $\rho$
& $0.0025$ & $0.0025$ \\
Gaussian smoothing $\sigma_{\mathrm{g}}$
& $4$ & $0$ \\
\midrule
State loss weight $\lambda_{\mathrm{state}}$
& \multicolumn{2}{c}{$0.10$} \\
State margin $\tau$
& \multicolumn{2}{c}{$0.50$} \\
Sampled patch pairs $|\mathcal{R}|$
& \multicolumn{2}{c}{$4{,}096$} \\
\bottomrule
\end{tabular}
\end{table}

\medskip
\noindent\textbf{Implementation details.}
For the selected DINOv3 ViT-L/16 backbone, images are resized
to $518\times518$, producing a $32\times32$ grid of
$1024$-dimensional patch tokens. The coefficient predictor
consists of Linear$(1024,64)$--GELU--Dropout$(0.1)$--Linear$(64,2)$.
Training uses AdamW for 10 epochs with learning rate $10^{-3}$,
weight decay $0.01$, batch size 2, gradient-norm clipping
at $1.0$, and horizontal and vertical flips.
All experiments are implemented in PyTorch and performed
on an NVIDIA RTX 6000 Ada GPU.

\section{Adaptation Design and State Space Analysis}
\label{sec:supp_state_analysis}

\noindent\textbf{Adaptation design ablation.}
To examine the design of SALA, Table~\ref{tab:supp_adaptation_ablation}
compares four feature adaptation mechanisms. All configurations share
the same frozen backbone, structural state formulation, prediction
function, DSA, Geometry Preservation, training settings, and source-only
evaluation protocol. Only the feature correction
$\widehat{\mathbf{z}}_i-\mathbf{z}_i$ is changed:
a) \textbf{Matched MLP}, where a residual MLP with the same number
of trainable parameters as SALT performs unrestricted adaptation,
$\widehat{\mathbf{z}}_i = \mathbf{z}_i+
\alpha f_{\phi}(\mathbf{z}_i)$;
b) \textbf{State rank-1}, where adaptation follows the direct
direction between the positive and negative states,
$\widehat{\mathbf{z}}_i = \mathbf{z}_i+
\alpha c_i\mathbf{u}_{\mathrm{D}}$,
with a scalar coefficient $c_i$ predicted by $g_{\theta}$
with a single output, and
$\mathbf{u}_{\mathrm{D}}\propto
\bar{\mathbf{s}}_{\mathrm{P}}-\bar{\mathbf{s}}_{\mathrm{N}}$,
where $\bar{\mathbf{s}}=\mathbf{s}/\|\mathbf{s}\|_2$;
c) \textbf{Free rank-2}, where a learned basis
$\mathbf{V}\in\mathbb{R}^{d\times 2}$ is independent
of the states,
$\widehat{\mathbf{z}}_i =
\mathbf{z}_i+\alpha\mathbf{V}\mathbf{c}_i$; and
d) \textbf{SALT}, where the rank-2 basis is constructed
from the prediction states,
$\widehat{\mathbf{z}}_i =
\mathbf{z}_i+\alpha\mathbf{U}(\mathbf{s}_{\mathrm{N}},
\mathbf{s}_{\mathrm{P}})\mathbf{c}_i$.
This comparison examines whether state-derived adaptation
improves localization compared with unrestricted and
independently learned low-rank corrections.

\begin{table}[t]
\centering
\caption{Adaptation design ablation (\%, mean$\pm$standard
deviation over three seeds). All configurations share the
structural states, DSA, and Geometry Preservation, and
differ only in the feature correction. Best results are in bold.}
\label{tab:supp_adaptation_ablation}

\scriptsize
\setlength{\tabcolsep}{2.0pt}
\renewcommand{\arraystretch}{1.02}

\resizebox{\columnwidth}{!}{%
\begin{tabular}{@{}lcccc@{}}
\toprule
\textbf{Configuration} &
\textbf{P-AUROC} &
\textbf{P-AP} &
\textbf{F1} &
\textbf{IoU} \\
\midrule

\multicolumn{5}{c}{\textbf{LEVIR-CD$\rightarrow$WHU}} \\
\midrule

Matched MLP
& $96.6\pm0.3$
& $84.2\pm0.8$
& $73.8\pm1.1$
& $58.5\pm1.3$ \\

State rank-1
& $\mathbf{98.2\pm0.1}$
& $87.1\pm0.4$
& $78.9\pm0.5$
& $65.1\pm0.7$ \\

Free rank-2
& $96.1\pm0.4$
& $82.9\pm0.7$
& $72.6\pm0.9$
& $57.1\pm1.1$ \\

SALT
& $98.1\pm0.1$
& $\mathbf{87.8\pm0.2}$
& $\mathbf{79.6\pm0.3}$
& $\mathbf{66.1\pm0.4}$ \\

\midrule
\multicolumn{5}{c}{\textbf{WHU$\rightarrow$LEVIR-CD}} \\
\midrule

Matched MLP
& $93.6\pm0.2$
& $43.3\pm0.6$
& $41.2\pm0.8$
& $25.9\pm0.7$ \\

State rank-1
& $94.5\pm0.1$
& $\mathbf{45.0\pm0.4}$
& $44.9\pm0.4$
& $28.9\pm0.5$ \\

Free rank-2
& $93.2\pm0.3$
& $42.6\pm0.5$
& $40.6\pm0.7$
& $25.5\pm0.5$ \\

SALT
& $\mathbf{94.8\pm0.2}$
& $44.4\pm1.0$
& $\mathbf{46.2\pm2.2}$
& $\mathbf{30.1\pm1.8}$ \\

\bottomrule
\end{tabular}%
}
\end{table}

\begin{figure}[t]
\centering
\begingroup
\setlength{\tabcolsep}{2pt}
\renewcommand{\arraystretch}{0}

\begin{tabular}{@{}cc@{}}

{\scriptsize\bfseries (a) Current image} &
{\scriptsize\bfseries (b) Ground truth} \\[1.5pt]

\includegraphics[width=.445\columnwidth]
{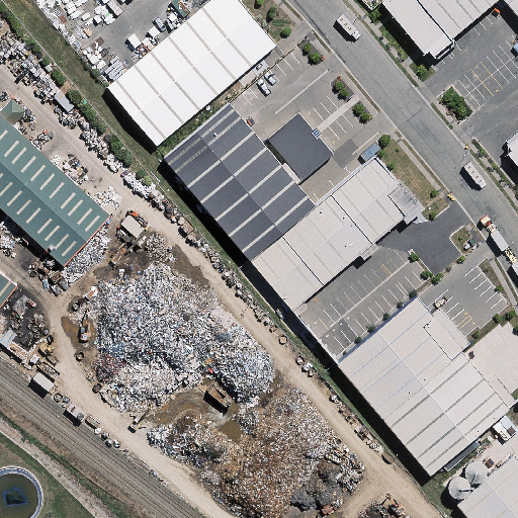} &
\includegraphics[width=.445\columnwidth]
{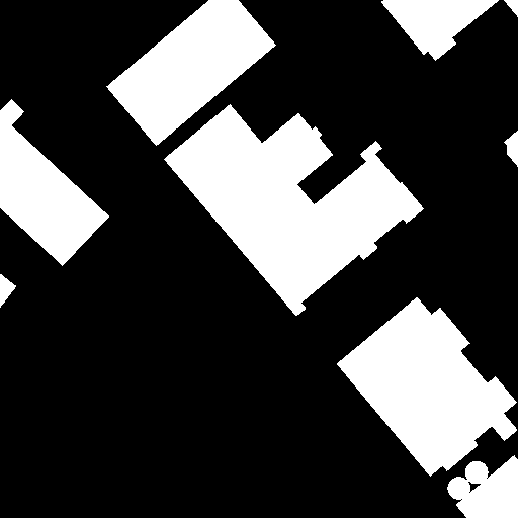} \\[3pt]

{\scriptsize\bfseries (c) Free rank-2
\textnormal{(IoU 77.3\%)}} &
{\scriptsize\bfseries (d) SALT
\textnormal{(IoU 82.9\%)}} \\[1.5pt]

\includegraphics[width=.445\columnwidth]
{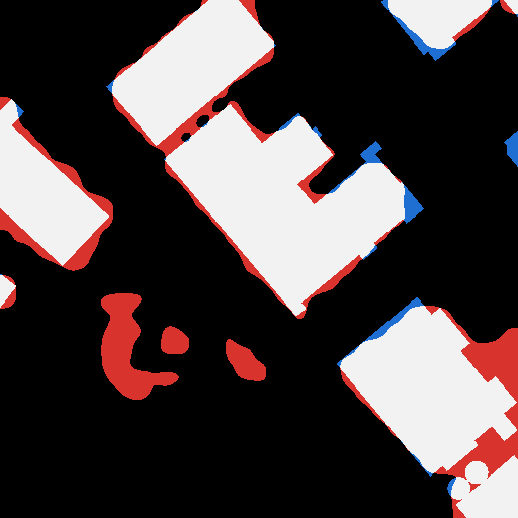} &
\includegraphics[width=.445\columnwidth]
{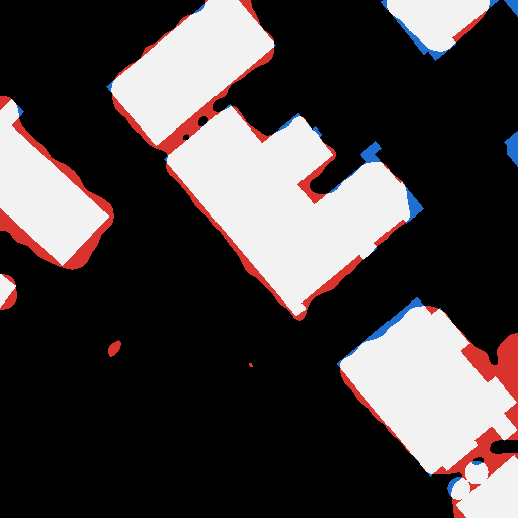}

\end{tabular}
\endgroup

\vspace{-1.5mm}

\caption{\textbf{Qualitative adaptation design comparison.}
Localization results on an unseen WHU image.
White indicates true positives, red false positives,
blue false negatives, and black true negatives.
Both methods recover the main building regions, while
SALT reduces spurious background predictions compared
with Free rank-2. IoU values correspond to the
illustrated image.}
\label{fig:supp_adaptation_qualitative}
\end{figure}

\medskip
\noindent\textbf{Discussion.}
Table~\ref{tab:supp_adaptation_ablation} supports two observations.
a) \textbf{State-aligned adaptation outperforms unrestricted
and generic low-rank corrections.} Both State rank-1 and
SALT outperform Matched MLP and Free rank-2 across all
pixel-level metrics and transfer directions, indicating that
the directions of feature correction matter beyond adaptation
capacity or low-rank constraints alone. Figure~\ref{fig:supp_adaptation_qualitative}
further illustrates this difference on an unseen WHU image,
where SALT reduces spurious background predictions compared
with Free rank-2 while preserving the main building regions,
improving image-specific IoU from $77.3\%$ to $82.9\%$.
b) \textbf{State rank-1 remains competitive with SALT},
achieving slightly higher P-AUROC in
LEVIR-CD$\rightarrow$WHU and P-AP in
WHU$\rightarrow$LEVIR-CD. SALT improves the remaining
metrics, suggesting modest, metric-dependent benefits
from the additional state-derived direction.

\medskip
\noindent\textbf{State separation and basis conditioning.}
Since SALA constructs its rank-2 adaptation space from the learned structural
states, their separation and conditioning are examined across the final
three-seed runs. Table~\ref{tab:supp_state_conditioning} reports the cosine
similarity between $\mathbf{s}_{\mathrm{N}}$ and $\mathbf{s}_{\mathrm{P}}$ and the remaining
positive-state component after Gram--Schmidt projection,
$\widetilde{\mathbf{s}}_{\mathrm{P}}
=\mathbf{s}_{\mathrm{P}}-(\mathbf{u}_{\mathrm{N}}^{\top}\mathbf{s}_{\mathrm{P}})\mathbf{u}_{\mathrm{N}}$.

\begin{table}[H]
\centering
\caption{State separation and SALA basis conditioning.}
\label{tab:supp_state_conditioning}

\footnotesize
\setlength{\tabcolsep}{3.5pt}
\renewcommand{\arraystretch}{1.05}

\begin{tabular}{@{}lcc@{}}
\toprule
\textbf{Direction} &
$\cos(\mathbf{s}_{\mathrm{N}},\mathbf{s}_{\mathrm{P}})$ &
$\|\widetilde{\mathbf{s}}_{\mathrm{P}}\|_2$ \\
\midrule

LEVIR-CD$\rightarrow$WHU
& $-0.8817\pm0.0034$
& $0.1978\pm0.0191$ \\

WHU$\rightarrow$LEVIR-CD
& $-0.9181\pm0.0004$
& $0.2166\pm0.0256$ \\

\bottomrule
\end{tabular}
\end{table}

From Table~\ref{tab:supp_state_conditioning}, the strongly negative cosine
similarities show that the two states encode opposing structural evidence.
At the same time, the orthogonal component remains nonzero in both directions,
showing that the second SALA direction remains defined and that no collapse
of the rank-2 adaptation space is observed in the evaluated runs.

\begin{figure*}[t]
    \centering
    \begin{minipage}[t]{0.49\textwidth}
        \centering
        \includegraphics[width=\linewidth]{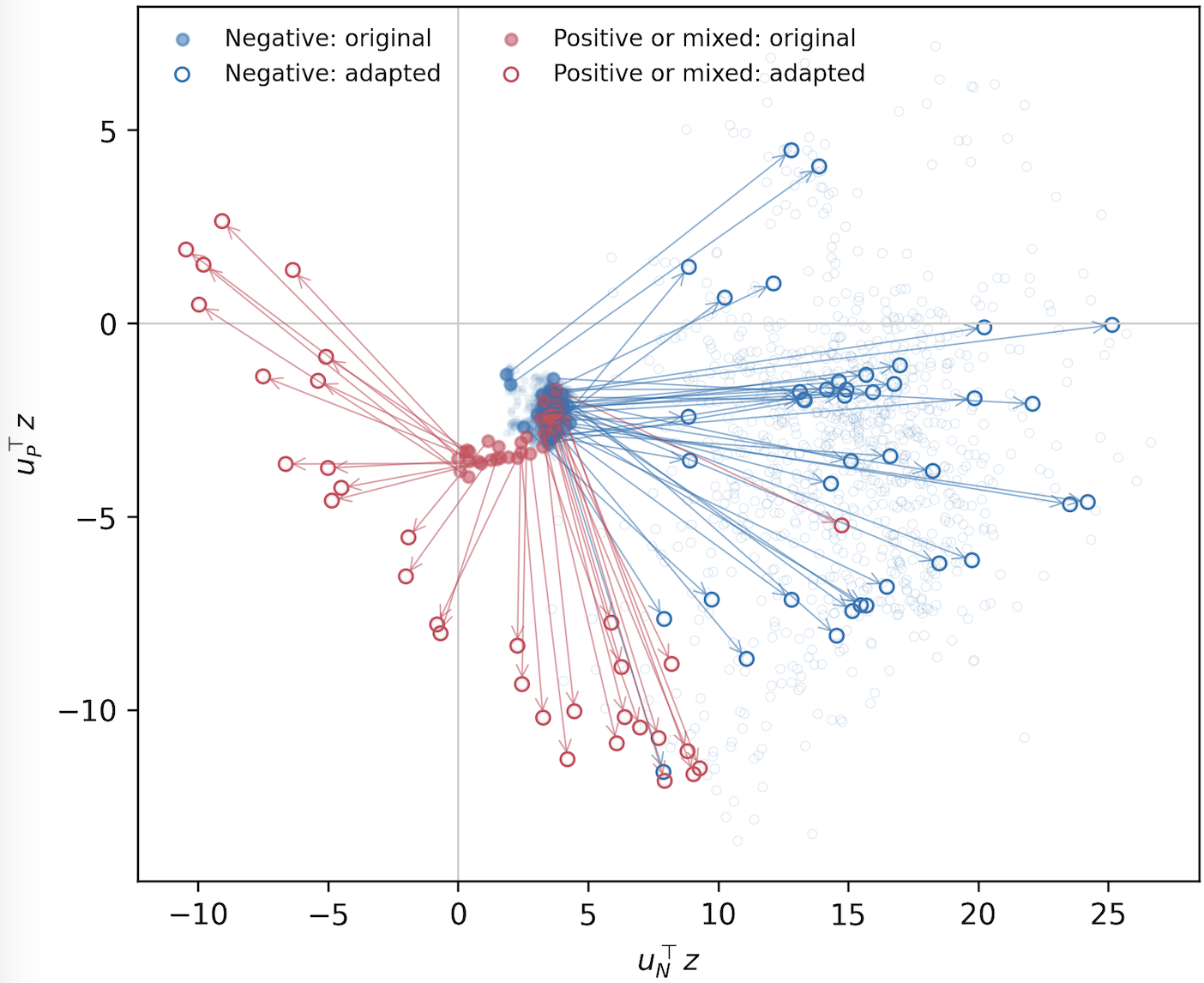}\\[-1mm]
        \small (a) LEVIR-CD$\rightarrow$WHU
    \end{minipage}
    \hfill
    \begin{minipage}[t]{0.49\textwidth}
        \centering
        \includegraphics[width=\linewidth]{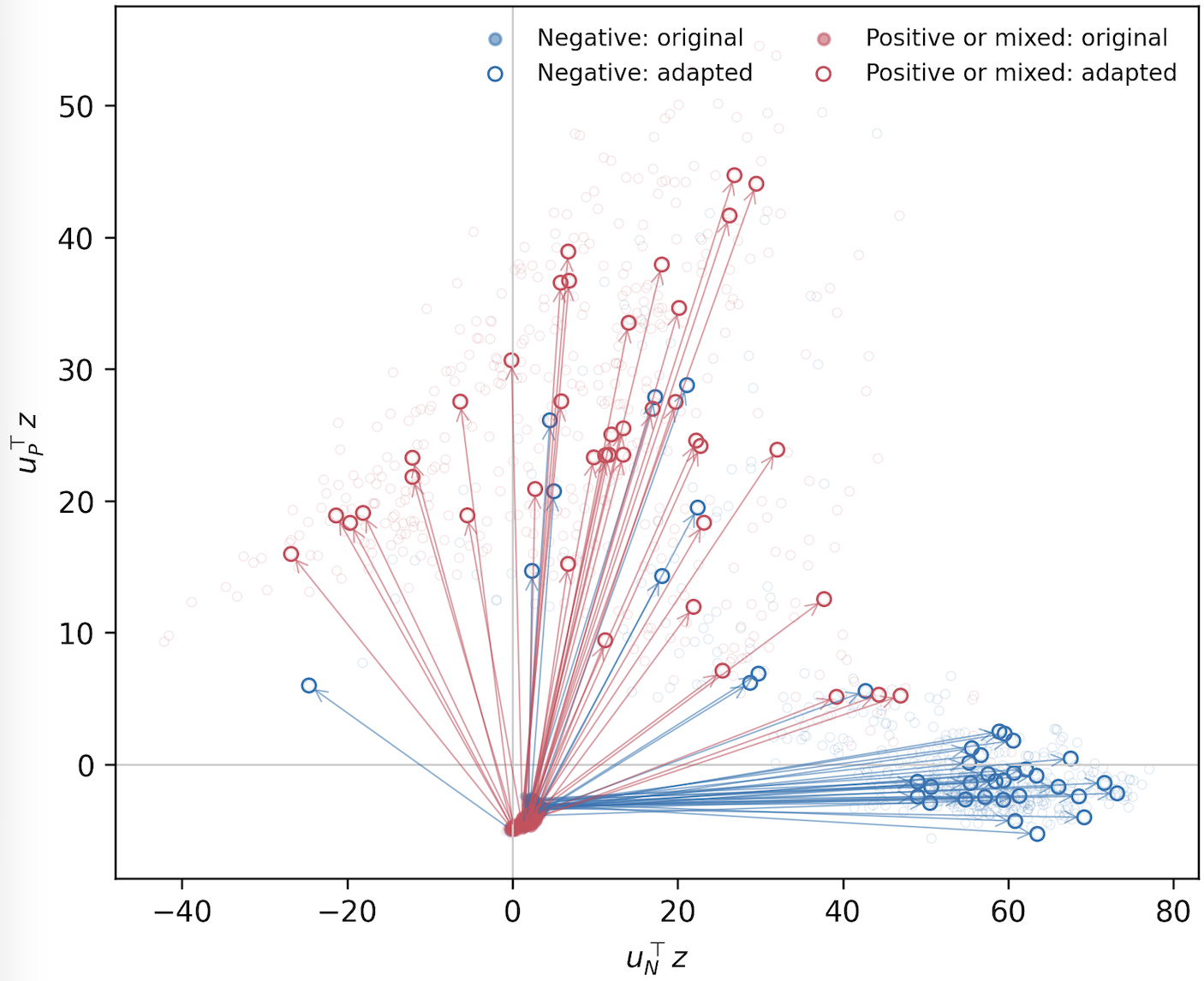}\\[-1mm]
        \small (b) WHU$\rightarrow$LEVIR-CD
    \end{minipage}

    \caption{Feature movement in the learned SALA space.
    Filled and open markers denote original and adapted features,
    respectively, while arrows indicate the corresponding corrections.
    Blue denotes negative patches and red positive or mixed patches.}
    \label{fig:supp_state_space}
\end{figure*}

\medskip
\noindent\textbf{Feature movement in the state space.}
To further examine the effect of SALA, target patch features before and after
adaptation are projected directly onto the learned basis as
$(\mathbf{u}_{\mathrm{N}}^{\top}\mathbf{z},
\mathbf{u}_{\mathrm{P}}^{\top}\mathbf{z})$.
Hence, the visualization corresponds directly to the state-defined space in
which SALA performs its feature corrections. Target annotations are used only
to distinguish negative from positive or mixed patches for visualization.
From Fig.~\ref{fig:supp_state_space}, SALA produces patch-dependent,
class-related corrections within the learned state space.
LEVIR-CD$\rightarrow$WHU exhibits clearer separation between negative and
positive-content patches, while WHU$\rightarrow$LEVIR-CD shows greater overlap
and more heterogeneous feature movement, consistent with the lower
localization performance in this transfer direction. The effects of DSA
and Geometry Preservation are examined quantitatively in Supplementary
Sections~\ref{sec:supp_dsa} and \ref{sec:supp_geometry}, respectively.

\section{Directional State Alignment Analysis}
\label{sec:supp_dsa}

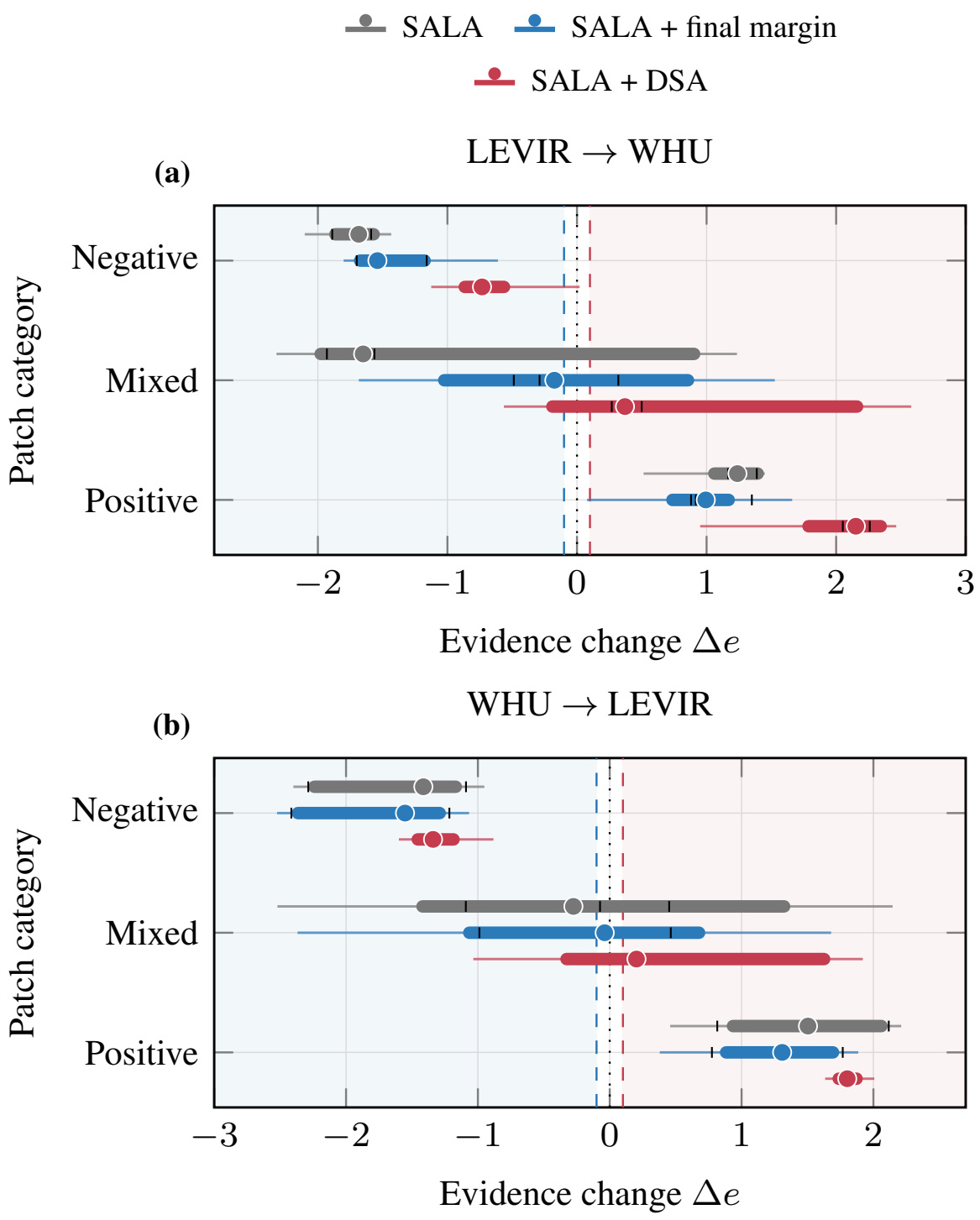
\begin{figure}[t]
\centering
\begingroup
\definecolor{saltgray}{HTML}{777777}
\definecolor{marginblue}{HTML}{2B7BBA}
\definecolor{dsared}{HTML}{C43C4E}
\definecolor{normalshade}{HTML}{2474B5}
\definecolor{positiveshade}{HTML}{C43C4E}

\newcommand{\EvidenceGlyph}[7]{%
  \draw[#1, line width=0.55pt, opacity=0.72]
    (axis cs:#3,#2) -- (axis cs:#7,#2);
  \draw[#1, line width=2.8pt, line cap=round]
    (axis cs:#4,#2) -- (axis cs:#6,#2);
  \addplot[only marks, mark=*, mark size=2.05pt,
    draw=white, line width=0.35pt, fill=#1]
    coordinates {(#5,#2)};
}

\newcommand{\SeedTicks}[4]{%
  \draw[black, line width=0.4pt]
    (axis cs:#2,{#1-0.045}) -- (axis cs:#2,{#1+0.045});
  \draw[black, line width=0.4pt]
    (axis cs:#3,{#1-0.045}) -- (axis cs:#3,{#1+0.045});
  \draw[black, line width=0.4pt]
    (axis cs:#4,{#1-0.045}) -- (axis cs:#4,{#1+0.045});
}

\begin{tikzpicture}
\begin{groupplot}[
  group style={
    group size=1 by 2,
    vertical sep=1.60cm
  },
  width=0.90\columnwidth,
  height=0.52\columnwidth,
  tick label style={font=\footnotesize},
  label style={font=\footnotesize},
  title style={font=\footnotesize},
  grid=major,
  grid style={draw=black!12, line width=0.25pt},
  axis line style={line width=0.55pt},
  tick style={line width=0.45pt},
]

\nextgroupplot[
  title={LEVIR $\rightarrow$ WHU},
  xmin=-2.8, xmax=3.0,
  ymin=-0.46, ymax=2.46,
  xlabel={Evidence change $\Delta e$},
  ylabel={Patch category},
  ytick={0,1,2},
  yticklabels={Positive,Mixed,Negative},
]
\path[fill=normalshade, fill opacity=0.055]
  (axis cs:-2.8,-0.46) rectangle (axis cs:-0.1,2.46);
\path[fill=positiveshade, fill opacity=0.055]
  (axis cs:0.1,-0.46) rectangle (axis cs:3.0,2.46);
\draw[normalshade, dashed, line width=0.45pt]
  (axis cs:-0.1,-0.46) -- (axis cs:-0.1,2.46);
\draw[dsared, dashed, line width=0.45pt]
  (axis cs:0.1,-0.46) -- (axis cs:0.1,2.46);
\draw[black, dotted, line width=0.45pt]
  (axis cs:0,-0.46) -- (axis cs:0,2.46);

\EvidenceGlyph{saltgray}{0.22}{0.513785}{1.059820}{1.238143}{1.394832}{1.449162}
\SeedTicks{0.22}{1.387621}{1.300427}{1.166176}

\EvidenceGlyph{marginblue}{0.00}{0.077146}{0.736448}{0.993931}{1.170897}{1.660533}
\SeedTicks{0.00}{0.880497}{1.349488}{0.916820}

\EvidenceGlyph{dsared}{-0.22}{0.950570}{1.786941}{2.151836}{2.344114}{2.463643}
\SeedTicks{-0.22}{2.259444}{2.051569}{2.091201}

\EvidenceGlyph{saltgray}{1.22}{-2.319475}{-1.978332}{-1.650645}{0.903930}{1.234520}
\SeedTicks{1.22}{-1.563113}{-1.622898}{-1.930292}

\EvidenceGlyph{marginblue}{1.00}{-1.684232}{-1.024764}{-0.174804}{0.856684}{1.526787}
\SeedTicks{1.00}{-0.487784}{0.319066}{-0.288879}

\EvidenceGlyph{dsared}{0.78}{-0.564184}{-0.190226}{0.370036}{2.161754}{2.580074}
\SeedTicks{0.78}{0.499275}{0.268259}{0.347112}

\EvidenceGlyph{saltgray}{2.22}{-2.101339}{-1.865081}{-1.684530}{-1.569472}{-1.433832}
\SeedTicks{2.22}{-1.627513}{-1.588919}{-1.889019}

\EvidenceGlyph{marginblue}{2.00}{-1.800938}{-1.673942}{-1.540263}{-1.176039}{-0.610269}
\SeedTicks{2.00}{-1.573057}{-1.159953}{-1.700137}

\EvidenceGlyph{dsared}{1.78}{-1.125192}{-0.866217}{-0.733176}{-0.563406}{0.020256}
\SeedTicks{1.78}{-0.731817}{-0.716656}{-0.753036}

\nextgroupplot[
  title={WHU $\rightarrow$ LEVIR},
  xmin=-3.0, xmax=2.7,
  ymin=-0.46, ymax=2.46,
  xlabel={Evidence change $\Delta e$},
  ylabel={Patch category},
  ytick={0,1,2},
  yticklabels={Positive,Mixed,Negative},
]
\path[fill=normalshade, fill opacity=0.055]
  (axis cs:-3.0,-0.46) rectangle (axis cs:-0.1,2.46);
\path[fill=positiveshade, fill opacity=0.055]
  (axis cs:0.1,-0.46) rectangle (axis cs:2.7,2.46);
\draw[normalshade, dashed, line width=0.45pt]
  (axis cs:-0.1,-0.46) -- (axis cs:-0.1,2.46);
\draw[dsared, dashed, line width=0.45pt]
  (axis cs:0.1,-0.46) -- (axis cs:0.1,2.46);
\draw[black, dotted, line width=0.45pt]
  (axis cs:0,-0.46) -- (axis cs:0,2.46);

\EvidenceGlyph{saltgray}{0.22}{0.458598}{0.934826}{1.503725}{2.061382}{2.210813}
\SeedTicks{0.22}{1.507070}{2.116349}{0.815638}

\EvidenceGlyph{marginblue}{0.00}{0.379279}{0.882711}{1.307858}{1.693537}{1.884888}
\SeedTicks{0.00}{1.320190}{1.767364}{0.775137}

\EvidenceGlyph{dsared}{-0.22}{1.634013}{1.736795}{1.801625}{1.872617}{2.006666}
\SeedTicks{-0.22}{1.808591}{1.783412}{1.811955}

\EvidenceGlyph{saltgray}{1.22}{-2.520411}{-1.420222}{-0.275407}{1.322947}{2.145587}
\SeedTicks{1.22}{-0.073482}{0.450963}{-1.091525}

\EvidenceGlyph{marginblue}{1.00}{-2.367856}{-1.065224}{-0.038221}{0.677885}{1.679322}
\SeedTicks{1.00}{-0.022550}{0.463147}{-0.988575}

\EvidenceGlyph{dsared}{0.78}{-1.034539}{-0.327644}{0.203066}{1.626147}{1.919812}
\SeedTicks{0.78}{0.223366}{0.208696}{0.173660}

\EvidenceGlyph{saltgray}{2.22}{-2.399491}{-2.240083}{-1.413684}{-1.167658}{-0.951090}
\SeedTicks{2.22}{-1.412278}{-1.090157}{-2.286517}

\EvidenceGlyph{marginblue}{2.00}{-2.523636}{-2.365552}{-1.552682}{-1.289658}{-1.066860}
\SeedTicks{2.00}{-1.553751}{-1.216241}{-2.414881}

\EvidenceGlyph{dsared}{1.78}{-1.599301}{-1.456050}{-1.340012}{-1.184790}{-0.882095}
\SeedTicks{1.78}{-1.337109}{-1.342359}{-1.340578}

\end{groupplot}

\node[anchor=south, font=\scriptsize] at
  ($(group c1r1.north)+(0,1.15cm)$) {%
  \tikz[baseline=-0.55ex]
    \draw[saltgray,line width=1.2pt]
    (0,0)--(0.32,0)
    node[pos=.5,circle,fill=saltgray,inner sep=1.0pt]{};
  \hspace{0.18em}SALA\hspace{0.65em}
  \tikz[baseline=-0.55ex]
    \draw[marginblue,line width=1.2pt]
    (0,0)--(0.32,0)
    node[pos=.5,circle,fill=marginblue,inner sep=1.0pt]{};
  \hspace{0.18em}SALA + final margin};

\node[anchor=south, font=\scriptsize] at
  ($(group c1r1.north)+(0,0.78cm)$) {%
  \tikz[baseline=-0.55ex]
    \draw[dsared,line width=1.2pt]
    (0,0)--(0.32,0)
    node[pos=.5,circle,fill=dsared,inner sep=1.0pt]{};
  \hspace{0.18em}SALA + DSA};

\node[anchor=south east, font=\bfseries\scriptsize] at
  ($(group c1r1.north west)+(-0.06cm,0.02cm)$) {(a)};

\node[anchor=south east, font=\bfseries\scriptsize] at
  ($(group c1r2.north west)+(-0.06cm,0.02cm)$) {(b)};

\end{tikzpicture}

\endgroup
\caption{\textbf{Directional State Alignment analysis.}
(a,b) Changes in structural evidence for negative,
mixed, and positive patches on the LEVIR and WHU
source validation sets. Thick bars denote interquartile
ranges and thin lines denote the 5th to 95th percentile
ranges. Circles indicate pooled medians, black ticks
indicate seed medians, and dashed lines show the
directional margins.}
\label{fig:supp_dsa_analysis}
\end{figure}

To examine the effect of DSA, Fig.~\ref{fig:supp_dsa_analysis}
compares plain SALA, SALA with a conventional margin on the
final structural evidence, and SALA with DSA. Unlike the
conventional margin, which supervises the final evidence,
DSA encourages positive or negative changes in evidence,
$\Delta e_i=e(\widehat{\mathbf{z}}_i)-e(\mathbf{z}_i)$.
For this analysis, a common margin of $m=0.1$ is used for
both source datasets, while the final WHU-source configuration
uses $m=0.2$, as reported in Table~\ref{tab:supp_final_config}. The analysis uses source
validation patches to examine these changes without
requiring target annotations. Patches are grouped as
negative, mixed, or positive based on their positive
pixel occupancy. With plain SALA, mixed patches receive
predominantly negative median corrections. Adding DSA
shifts the mean of the three seed medians from $-1.705$
to $+0.372$ on LEVIR and from $-0.238$ to $+0.202$
on WHU. In comparison, the conventional margin produces
values of $-0.153$ and $-0.183$, respectively.
DSA also produces stronger positive corrections for
positive patches, while negative patches retain
predominantly negative corrections. These results show
that DSA changes how SALA adjusts structural evidence,
particularly for patches containing both positive
and negative regions.

\section{Geometry Preservation Analysis}
\label{sec:supp_geometry}

\begin{figure}[t]
\centering
\begingroup
\definecolor{geoblue}{HTML}{2B7BBA}

\begin{tikzpicture}
\begin{groupplot}[
  group style={
    group size=1 by 2,
    vertical sep=2.45cm
  },
  width=0.94\columnwidth,
  height=0.37\columnwidth,
  xmin=0, xmax=46,
  ymin=0.55, ymax=2.45,
  ytick={1,2},
  yticklabels={
    $W\rightarrow L$,
    $L\rightarrow W$
  },
  xmajorgrids=true,
  grid style={
    draw=black!12,
    line width=0.3pt
  },
  xtick={0,10,20,30,40},
  xlabel={Reduction compared with no Geometry (\%)},
  tick label style={font=\scriptsize},
  label style={font=\scriptsize},
  title style={font=\footnotesize},
  tick style={line width=0.4pt},
  axis line style={line width=0.5pt},
]

\nextgroupplot[
  title={(a) Pairwise cosine distortion $D$}
]

\path[fill=geoblue!85]
  (axis cs:0,1.78) rectangle
  (axis cs:21.2446,2.22);

\path[fill=geoblue!85]
  (axis cs:0,0.78) rectangle
  (axis cs:5.4954,1.22);

\node[anchor=west, font=\footnotesize]
  at (axis cs:22.1,2) {21.2\%};

\node[anchor=west, font=\footnotesize]
  at (axis cs:6.4,1) {5.5\%};

\nextgroupplot[
  title={(b) Relative feature correction $M$}
]

\path[fill=geoblue!85]
  (axis cs:0,1.78) rectangle
  (axis cs:35.6309,2.22);

\path[fill=geoblue!85]
  (axis cs:0,0.78) rectangle
  (axis cs:22.6433,1.22);

\node[anchor=west, font=\footnotesize]
  at (axis cs:36.4,2) {35.6\%};

\node[anchor=west, font=\footnotesize]
  at (axis cs:23.5,1) {22.6\%};

\end{groupplot}
\end{tikzpicture}
\endgroup

\caption{\textbf{Geometry Preservation analysis on
unseen target images.}
Percentage reduction in (a) pairwise cosine
similarity changes and (b) relative feature
correction magnitude when Geometry Preservation
is added to SALA with DSA.
$L$ and $W$ denote LEVIR and WHU.
Reductions are calculated from the means
over three seeds.}
\label{fig:supp_geometry}
\end{figure}
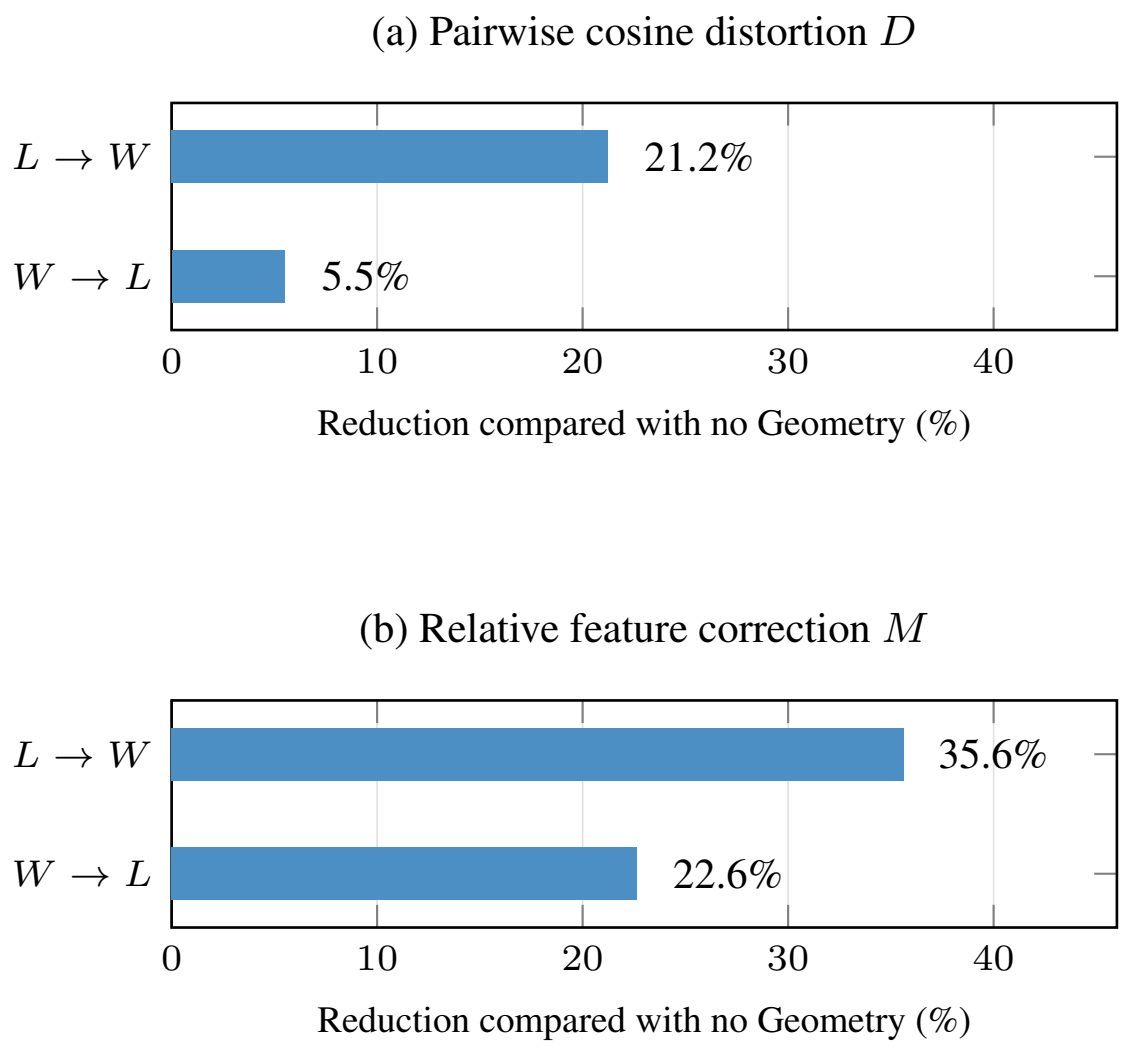

To examine the effect of Geometry Preservation,
Fig.~\ref{fig:supp_geometry} compares SALA with DSA
against the complete SALT model on unseen target images.
The analysis measures the change in pairwise cosine
similarity between original and adapted DINOv3 patch
features ($D$), together with the relative magnitude
of feature corrections ($M$). For $D$, both variants
use the same 4,096 sampled patch pairs per image.
Across three seeds, Geometry Preservation reduces
pairwise similarity changes by 21.2\% for
LEVIR$\rightarrow$WHU and 5.5\% for
WHU$\rightarrow$LEVIR. Relative feature correction
magnitudes decrease by 35.6\% and 22.6\%, respectively. These reductions show that Geometry Preservation
better maintains pretrained feature relationships.
Its effect on localization remains transfer- and
metric-dependent, as shown in Table~\ref{tab:component_ablation} of the main
paper.

\end{document}